\documentclass[twocolumn,10pt]{asme2ej}

\usepackage[T1]{fontenc}
\usepackage[noadjust]{cite}
\renewcommand{\citedash}{--}

\usepackage{amsmath}
\usepackage{yhmath}
\usepackage{mathtools} 
\usepackage{amssymb} 
\usepackage{sidecap}
\usepackage{graphicx}
\usepackage{subcaption}
\usepackage{multirow}
\usepackage{epsfig}
\usepackage{psfrag}
\usepackage{caption}
\usepackage[super]{nth} 
\usepackage[normalem]{ulem} 
\usepackage{color}
\usepackage{xcolor} 
\usepackage{booktabs} 
\usepackage{siunitx} 
\usepackage{booktabs} 
\usepackage{colortbl} 
\usepackage{textcomp} 
\usepackage{tikz}
\usepackage{bm}
\usepackage[toc,page]{appendix}
\usepackage[noadjust]{cite} 
\usepackage{fancyhdr}

\usepackage{wasysym}  

\newcommand{\RNum}[1]{\uppercase\expandafter{\romannumeral #1\relax}} 

\newcommand{\realfield}[1]{\hbox{I \kern -.4em R}^{#1}}
\newcommand {\mb}[1]{\mathbf{#1}}
\newcommand {\bs}[1]{\boldsymbol{#1}}
\newcommand{\uvec}[1]{\hat{\mathbf{#1}}}
\newcommand{\T}{^{\mathrm{T}}}  

\definecolor{LightGray}{gray}{0.9}
\newcolumntype{a}{>{\columncolor{LightGray}}l}
\newcommand*\circled[1]{\tikz[baseline=(char.base)]{\node[circle,minimum size=7pt,draw=black,inner sep=0.5pt](char){\scriptsize #1};}}

\usepackage{balance}

\newif\ifTrackChanges   
\TrackChangesfalse 

\definecolor{deep-red}{RGB}{192, 0, 0}
\definecolor{deep-purple}{RGB}{120, 0, 170}
\definecolor{good-green}{RGB}{0,175,0}
\definecolor{purple}{RGB}{210, 0, 210}
\definecolor{alizarin}{rgb}{0.82, 0.1, 0.26}

\usepackage[textsize=scriptsize, bordercolor=white,backgroundcolor=gray!30,linecolor=black,colorinlistoftodos]{todonotes}  
\usepackage[normalem]{ulem} 
\usepackage{soul} 

\usepackage[mathscr]{euscript}
\ifTrackChanges
    \newcommand{\cut}[1]{{\color{gray}{#1}}}
    \newcommand{\remind}[2]{{[\colorbox{cyan}{#1}]}{\color{blue}{#2}}}  
    \newcommand{\corrlab}[2]{{[\colorbox{yellow}{#1}]~}{\color{red}{#2}}} 
    \newcommand{\cusst}[1]{{\st{#1}}}  
\else
     \newcommand{\cut}[1]{}
     \newcommand{\remind}[2]{#2}
    \newcommand{\corrlab}[2]{#2}
    
    \newcommand{\cusst}[1]{} 
\fi

\title{Design Considerations and Robustness to Parameter Uncertainty in Wire-Wrapped Cam Mechanisms}
\author{Garrison L.H. Johnston
    \affiliation{
	Dept. of Mechanical Engineering\\
	Vanderbilt University\\
	Nashville, Tennessee 37235\\
    Garrison.L.Johnston@vanderbilt.edu
    }	
}

\author{Andrew L. Orekhov
    \affiliation{
    Dept. of Mechanical Engineering\\
	Vanderbilt University\\
	Nashville, Tennessee 37235\\
    Andrew.Orekhov@vanderbilt.edu
    }
}

\author{Nabil Simaan\thanks{Address all correspondence for other issues to this author.}
    \affiliation{
    Dept. of Mechanical Engineering\\
    Vanderbilt University\\
    Nashville, Tennessee 37235\\
    Nabil.Simaan@vanderbilt.edu
    }
}

\graphicspath{{figures/pdf/}}

\makeatletter
\let\NAT@parse\undefined
\makeatother
\usepackage{hyperref}  
\usepackage{cancel}
\usepackage{cleveref}

\begin{document}


\pagestyle{fancyplain}
\fancyhf{} 
\fancyhead[L]{ASME Journal of Mechanisms and Robotics. Accepted Version}
\fancyhead[R]{DOI: \href{ https://doi.org/10.1115/1.4056600}{10.1115/1.4056600}}
\renewcommand{\headrulewidth}{0pt} 
\fancyfoot[R]{\thepage}
\fancyfoot[C]{Simaan}
\fancyfoot[L]{JMR-22-1290}

\captionsetup[table]{skip=-5pt,font=small}
\captionsetup{skip=0pt,font=small}
\maketitle
\begin{abstract}
\it Collaborative robots must simultaneously be safe enough to operate in close proximity to human operators and powerful enough to assist users in industrial tasks such as lifting heavy equipment. The requirement for safety necessitates that collaborative robots are designed with low-powered actuators. However, some industrial tasks may require the robot to have high payload capacity and/or long reach. For collaborative robot designs to be successful, they must find ways of addressing these conflicting design requirements. One promising strategy for navigating this tradeoff is through the use of static balancing mechanisms to offset the robot's self weight, thus enabling the selection of lower-powered actuators. In this paper, we introduce a novel, 2 degree of freedom static balancing mechanism based on spring-loaded, wire-wrapped cams. We also present an optimization-based cam design method that guarantees the cams stay convex, ensures the springs stay below their extensions limits, and minimizes sensitivity to unmodeled deviations from the nominal spring constant. Additionally, we present a model of the effect of friction between the wire and the cam. Lastly, we show experimentally that the torque generated by the cam mechanism matches the torque predicted in our modeling approach. Our results also suggest that the effects of wire-cam friction are significant for non-circular cams.
\end{abstract} 
\section{Introduction}
\subsection{Motivation}
\par The United States Bureau of Labor Statistics reports that in 2015, musculoskeletal disorders (MSDs), such as pinched nerves, herniated discs, and carpal tunnel syndrome, accounted for 31\% of all days-away-from-work cases reported to the agency \cite{MSD}. These work-related MSDs are caused by exerting strenuous and/or repetitive forces in awkward and non-ergonomic conditions \cite{Hales1996}. While full automation could eliminate the risk completely, the complexity of certain industrial tasks, such as equipment maintenance and repair, may necessitate a worker to be physically present at the worksite to control critical aspects of the task.
\par In an attempt to address this issue, there has been a recent push to develop \textit{in-situ collaborative robots} (ISCRs) that can help alleviate the physiological stresses of the aforementioned tasks by exerting the potentially harmful forces while being guided by the worker using physical human-robot interaction. Because of the worker's proximity to the robot, it is absolutely imperative that the robot be endowed with both active and passive measures of safety. Active safety measures, such as collision detection and avoidance algorithms, are safeguards against injury coded into the robot's control system. Passive safety measures, on the other hand, are features built into the robot's mechanical design that minimize the likelihood of trauma in the case of a collision between the robot and the worker. Designing a robot with as low powered actuators as possible can effectively increase the passive safety of the device, especially in the case of catastrophic control system failure. By removing the need for the actuators to lift the robot's self weight, static balancing mechanisms are a promising way of lowering actuator torque requirements and thereby achieving higher levels of passive safety \cite{Vermeulen2010}.
\subsection{Relevant Works and Summary of Limitations}
\par Static balancing mechanisms are typically realized using either counterbalancing masses (e.g., \cite{Laliberte1999,Whitney2014,Petrescu2018}), springs, or a combination of both \cite{Laliberte1999,Woo2019}. Spring-based static balancing mechanisms offer the advantage of lower overall mass and inertia compared to counterbalancing, making them an attractive option for safety, but come at the cost of higher design and modeling complexity. Spring-based static balancing methods have mostly fallen into three categories: 1) spring-loaded linkages \cite{Gopalswamy1992,Agrawal2005,Barents2011,Kim2013,Ahn2016,Chu2016,Takahashi2019}, 3) cams with spring-loaded followers \cite{Simionescu2000a,Wu2001,Koser2009,Lee2018,Buskiewicz2019}, and 3) and wire-wrapped cams \cite{harrington1951,carlson1975,Tidwell1994,Kilic2012,Schroeder2017,Spagnuolo2017,schroeder2018,Yigit2018,Fedorov2018,Kim2020,Candan2020,Qu2022}. Additionally, many simple balancing mechanisms are presented in \cite{Simionescu2000}.
\par It is possible to achieve exact static balancing of planar linkages by using zero-length springs and parallelogram linkages \cite{Morita2003,Kim2013,Lee2017}. While these design solutions may yield acceptable embodiments for certain applications, it may be preferable to consider designs that do not increase the footprint of the manipulator's linkages because of the restricted nature of some applications. A potential solution for this could be the more compact designs presented in \cite{Cho2010,Cho2012}. However, these designs are very complex and add many components such as springs, bevel gears, linear bearings, and linear slides into the linkages of the robot. This increases the mass and inertia of the robot, thus reducing its safety. Therefore there is interest in keeping as many balancing components as possible at the base of the robot.
\par This paper will focus on wire-wrapped cam methods. These mechanisms are typically realized with a wire that is fixed to the cam on one end, goes over an idler pulley, and is attached to a linear spring at the other end. When the cam rotates, it wraps the wire over the cam, inducing a torque on the cam from the resulting spring deflection. From a given spring constant, the cam profile can be specially designed to match a desired torque profile as a function of cam angle. Previous works in these areas have used either closed form equations (e.g., \cite{Kilic2012}) or graphical synthesis (e.g., \cite{Fedorov2018}) to determine the profile of the cam. \remind{R1-1}{These methods can provide a cam profile that will generate the desired torque profile exactly. However, they suffer from multiple drawbacks. For certain desired torque profiles, 1) the methods can return non-convex cams (which is needed for continued tangency between the wrapped wire and the cam), 2) there is no explicit guarantee the spring stay below the maximum allowable deflection and in many cases necessitate unrealistically large spring deflections, and 3) there is no consideration of how uncertainty in the spring stiffness may effect the balancing performance. Additionally, previous works have treated the torque on the cam as resulting from a point force applied at the wire-cam tangency point, ignoring the distributed friction and pressure between the rest of the wire and the cam}.
\par \corrlab{R1-1}{While it may be feasible to search for a set of input parameters that address the non-convexity problem of the closed-form solution, we believe this may not always be a feasible approach. The location and size of the idler is subject to space constraints. Similarly, the cam size is also governed by geometric constraints within a realistic design. The spring parameters that would satisfy cam convexity may also not be feasible when realistic spring design considerations are applied. Finally, this process of sampling the design space is essentially a time-consuming, ad-hoc design optimization process. In this paper, we offer an alternative approach that allows automatic consideration of geometric constraints, robustness to variation in spring parameters, and cam convexity. As such, our approach offers a design that considers realistic constraints while offering the best feasible approximation for the static balancing problem.}
\par Design optimization for static balancing has been used in the past to determine which static robot configuration to perfectly balance to achieve the best balancing across its workspace \cite{Mahalingam1986} and to select torsional springs to balance a medical robot \cite{Lessard2007}. It has also been used to determine the fourier coefficients of the cam dynamic equations \cite{Demeulenaere2005} and to optimize mechanism parameters for multi-DOF robots \cite{Saravanan2008}. To the best of our knowledge, design optimization has not been used to address the concerns with wire-wrapped cams listed in the previous paragraph.
\subsection{Contribution and Paper Organization}
\par The contribution of this paper relative to previous works is a cam design procedure that uses optimization to select modal coefficients for the cam shape and spring pre-extensions that minimize the difference between the actual cam torque and the desired cam torque. It also minimizes the sensitivity of the output torque to unanticipated changes in spring constant. The optimization problem is constrained to guarantee cam convexity and ensures the cam will not violate the maximum allowable spring deflections. As an additional contribution of this work, we model the effect of friction between the cam and the wire. This optimization routine is applied to a novel two degree-of-freedom (DOF) cam design. This two DOF system can be used for systems when the desired torque on one or both cams is a function of both cam angles. The major drawback of our method is that the torque on the cams is not guaranteed to exactly match the desired torque. However, for robotic applications, it is much more desirable to have a physically build-able cam system that approximately balances the static torques than a cam design that theoretically, exactly balances the static torques but violates the maximum allowable spring extensions and is not convex (i.e. does not work in practice).
\par This paper is organized as follows. In Section \ref{sec:alpha_gamma}, we present the design of a novel one DOF cam system and solve for the wire tangency point. In Section \ref{sec:spring_extensions}, we solve for the spring extensions for the one DOF cam design and extend the equations for a two DOF cam design. Later, in Section \ref{sec:one_dof_torque}, we calculate the torque on the one DOF cam system with and without ignoring frictional effects. In Section \ref{sec:2dof_torque}, we extend the equations for the two DOF cam design. Then in Section \ref{sec:modal}, we describe the modal basis chosen for the cam design and derive the conditions on the cam modal basis required for cam convexity. In Section \ref{sec:sens}, we derive a first order approximation of the sensitivity of the cam torque to unmodeled changes in spring torque. The optimization problem used to design the cams is formulated in Section \ref{sec:opt}. In Section \ref{sec:sim}, we present the results of two simulation case studies and in Section \ref{sec:experiment} our model for the torque on the cams is experimentally validated. Lastly, the results are discussed in Section \ref{sec:disc} and conclusions are drawn in Section \ref{sec:conclusion}.

\section{The Tangency Points in a Wire-Cam Mechanism}\label{sec:alpha_gamma}
\begin{figure*}[htbp]
    \centering
    \includegraphics[width=0.95\textwidth]{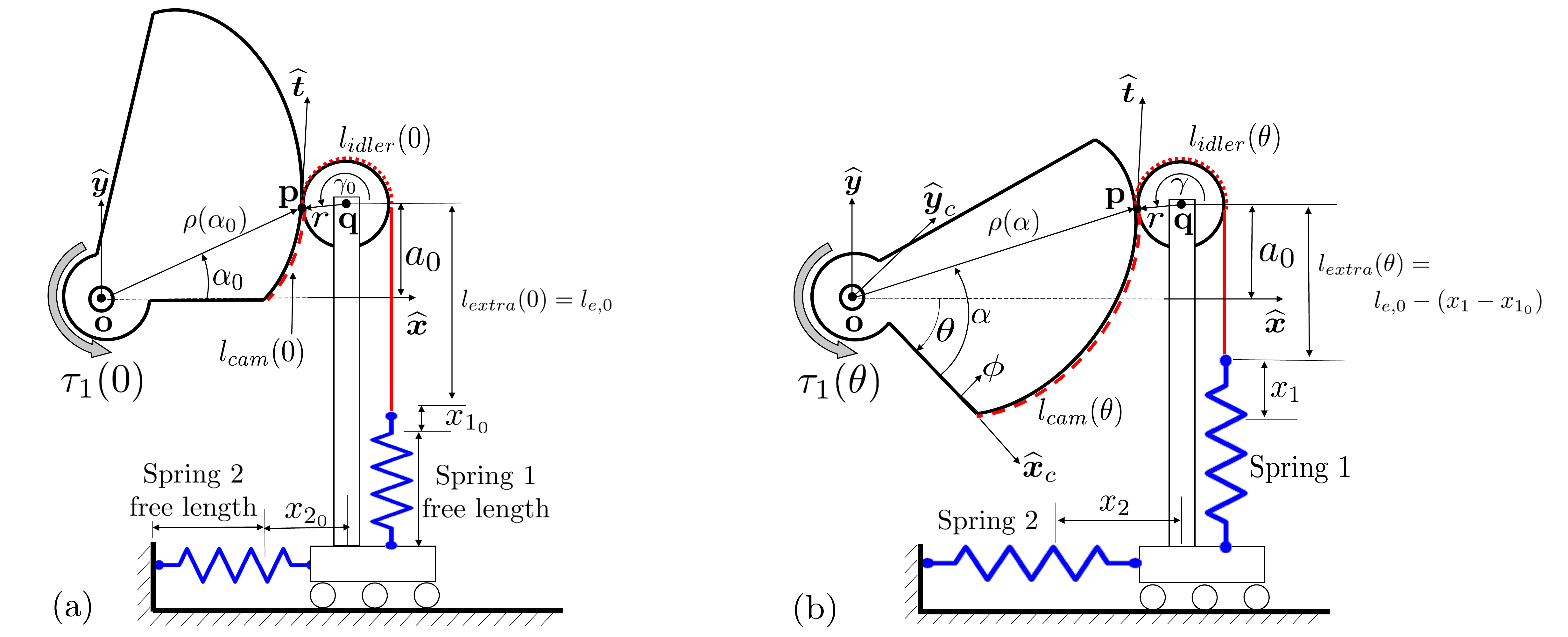}
    \caption{Cam design concept for a 1 DOF system: (a) $\theta = 0$ and (b) $\theta>0$}
    \label{fig:cam_1dof}
\end{figure*}
\par The calculation of spring extensions in wire-cam mechanisms requires knowledge of the tangency point between the idler and the cam. In this section, we present a numerical solution to finding the tangency point for a general wire-cam design. Specifically, we refer the reader to Fig.~\ref{fig:cam_1dof}, where a follower idler is pressed horizontally against the cam using one spring and the wire rope is tensioned using another spring. Figure \ref{fig:cam_1dof}(a) shows the system with initial spring preload and zero cam rotation.  Figure \ref{fig:cam_1dof}(b) shows the same system for a cam rotation $\theta > 0$.
\par The spring extensions $x_1$ and $x_2$ shown in \ref{fig:cam_1dof}(b) are functions of the angles $\alpha$ and $\gamma$. These two angles characterize the location of the tangency point $\mb{p}$ on the cam and idler, respectively.  To solve for these angles, we use the fact that the cam and idler are tangent to one another at $\mb{p}$. To do this we will calculate the unit vectors that are tangent to the cam $\hat{\mb{t}}(\alpha)$ and idler $\hat{\mb{t}}(\gamma)$ and set them equal.
\par The local tangents to the cam and idler, $\mb{t}(\alpha)$ and $\mb{t}(\gamma)$ are shown in Fig. ~\ref{fig:cam_1dof}.  These vectors are defined as:
\begin{equation}\label{Eq:tangents}
\mb{t}(\alpha)=\frac{d}{d\alpha}\mb{r}_{p/o}, \quad \mb{t}(\gamma)=\frac{d}{d\gamma}\mb{r}_{p/q}
\end{equation}
where the position vectors $\mb{r}_{p/o}$ and $\mb{r}_{p/q}$ are given by\footnote{We use $\mb{r}_{x/y}\triangleq\mb{x}-\mb{y}$ where $\mb{x}\in\realfield{3}$ and $\mb{y}\in\realfield{3}$ are points.}:
\begin{equation}\label{eq:rpo}
  \mb{r}_{p/o} = \rho(\alpha)\begin{bmatrix}
                               \cos(\alpha-\theta) \\
                               \sin(\alpha-\theta)\\
                               0
                             \end{bmatrix}, \qquad
  \mb{r}_{p/q} = r\begin{bmatrix}
                    \cos(\gamma) \\
                    \sin(\gamma)\\
                    0
                    \end{bmatrix}
\end{equation}
The explicit expressions for these tangent vectors are:
\begin{equation}\label{eq:tangent_partials}
 \mb{t}(\alpha)= \rho'(\alpha)\begin{bmatrix}
                                 \cos(\alpha-\theta) \\
                                 \sin(\alpha-\theta)\\
                                 0
                               \end{bmatrix} + \rho(\alpha)\begin{bmatrix}
                                                             -\sin(\alpha-\theta) \\
                                                             \cos(\alpha-\theta)\\
                                                             0
                                                           \end{bmatrix}
 \end{equation}
\begin{equation}\label{eq:tangent_gamma}
   \mb{t}(\gamma) = r \left[  -\sin(\gamma),~\cos(\gamma),~0\right]\T
\end{equation}
\noindent Normalizing these vectors using $\|\mb{t}(\alpha)\| = \sqrt{\rho(\alpha)^2+\rho'(\alpha)^2}$ and $\|\mb{t}(\gamma)\| = r$ gives the local tangent unit vectors:
\begin{equation}\label{eq:t}
  \hat{\mb{t}}(\alpha)= \frac{\mb{t}(\alpha)}{\sqrt{\rho(\alpha)^2+\rho'(\alpha)^2}} \qquad \hat{\mb{t}}(\gamma) = \frac{\mb{t}(\gamma)}{r}
\end{equation}
\noindent The local tangency constraint between the idler and cam is:
\begin{equation}\label{Eq:tangency_constraint}
\uvec{t}(\alpha)=-\uvec{t}(\gamma)
\end{equation}
The above constraint could be solved numerically, but can return more than one answer if $\alpha$ and $\gamma$ are not related through an added constraint. We choose the constraint that the $y$ coordinate of $\mb{r}_{p/o}$ is equal to sum of the y coordinates of $\mb{r}_{q/o}$ and $\mb{r}_{p/q}$. For example, if $\mb{r}_{q/o}\T\uvec{y}_0=a_0$ (i.e. the vertical offset of the idler in Fig. \ref{fig:cam_1dof}), then:
\begin{equation}\label{eq:y_contact}
  \rho(\alpha)\sin(\alpha-\theta) = r\sin(\gamma)+a_0
\end{equation}
\par \remind{R1-11}{Combining Eq.~\eqref{Eq:tangency_constraint} with Eq.~\eqref{eq:y_contact} allows us to formulate the following optimization problem:
\begin{equation}\label{eq:alpha_gamma}
\begin{aligned}
\min_{\alpha,\gamma}\quad & \|\uvec{t}(\alpha)+\uvec{t}(\gamma)\|^2 + \left(\rho(\alpha)\sin(\alpha-\theta) - r\sin(\gamma)-a_0\right)^2
\end{aligned}
\end{equation}

\noindent For any cam angle $\theta$, the Levenberg-Marquardt algorithm \cite{More1978_levenberg_marquardt} can be used to solve for $\alpha$ and $\gamma$ (e.g., using MATLAB's \textit{lsqnonlin} function).}
\section{Spring Extensions For Wire-Cam Mechanisms}\label{sec:spring_extensions}
Next, we will determine the spring deflections for design variants with one and two DOF.
\subsection{One DoF Wire-Wrapped Cams}\label{sec:1dof_extension}
 %
 \par Referring to Fig.~\ref{fig:cam_1dof}, the deflection in spring 1 can be solved using the assumption that the length $l$ of the wire remains constant. This wire length is shown in red in Fig. \ref{fig:cam_1dof} as the sum of three portions shown using dashed line $l_{cam}$, dotted line $l_{idler}$, and solid line $l_{extra}$.
\begin{equation}\label{eq:L}
  l = l_{cam}+l_{idler}+l_{extra}
\end{equation}
\noindent $l_{cam}$ can be found using the line integral along the cam from $0$ to $\alpha$. The profile of the cam is expressed using the polar function $\rho(\phi)$ where $\phi$ parameterizes the angle swept along the cam. This line integral can be written as:
\begin{equation}\label{eq:L_cam}
l_{cam}=\int_{0}^{\alpha}\sqrt{\rho(\phi)^2+\rho'(\phi)^2}d\phi
\end{equation}
\noindent Next, $l_{idler}$ can be expressed as:
\begin{equation}\label{eq:L_idler}
  l_{idler} = r\gamma
\end{equation}
\noindent where $r$ is the radius of the idler and $\gamma$ is the angle between the horizontal and the idler tangency point as shown in Fig. \ref{fig:cam_1dof}. Now, $l_{extra}$ can be found using:
\begin{equation}\label{eq:L_extra}
  l_{extra} = l_{e,0}-\left(x_1-x_{1_0}\right)
\end{equation}
\noindent In this equation, $l_{e,0}$ is the length of wire between spring 1 when $\theta = 0$, $x_1$ is the total extension of spring 1, and $x_{1_0}$ is the extension of spring 1 when $\theta = 0$ (i.e. the pre-extension of spring 1). Because the wire is inextensible, the change in string length between when $\theta > 0$ and $\theta = 0$ (i.e., $\Delta l\triangleq l(\theta) - l(0)$) must be zero:
%
\begin{equation}\label{eq:delta_L2}
  \Delta l = \Delta l_{cam}(\theta)+ \Delta l_{idler}(\theta)+ \Delta l_{extra}(\theta) =0
\end{equation}
where the length change terms in Eq.~\eqref{eq:delta_L2} are measured with respect to the respective lengths when $\theta=0$.

\par  Plugging Eqs. (\ref{eq:L_cam}-\ref{eq:L_extra}) into Eq. \eqref{eq:delta_L2}  and solving for $x_1$ gives:
%
\begin{equation}\label{eq:x1}
  x_1 = \int_{\alpha_0}^{\alpha}\sqrt{\rho(\phi)^2+\rho'(\phi)^2}d\phi+r(\gamma-\gamma_0) + x_{1_0}
\end{equation}
\par The vector $\mb{r}_{q/o}\triangleq \mb{q}-\mb{o}$ is calculated as:
\begin{equation}\label{eq:rqo1}
  \mb{r}_{q/o} = \begin{bmatrix}
                   \rho(\alpha)\cos(\alpha-\theta)-r\cos(\gamma) \\ a_0 \\ 0
                 \end{bmatrix}
\end{equation}
\par The total extension of spring 2 can be found using its pre-extension $x_{2_0}$ and the change in the $x$ component of $\mb{r}_{q/o}$:
\begin{equation}\label{eq:x2}
\begin{aligned}
  x_2 =& \rho(\alpha)\cos(\alpha-\theta)- r\cos(\gamma) \\
  &-(\rho(\alpha_0)\cos(\alpha_0) - r\cos(\gamma_0))+x_{2_0}
\end{aligned}
\end{equation}

\subsection{Two DoF Wire-Wrapped Cams}\label{sec:2dof_deflection}
\begin{figure}[htbp]
    \centering
    \includegraphics[trim={0.5cm 0 0.0cm 0cm}, clip,width=0.99\columnwidth]{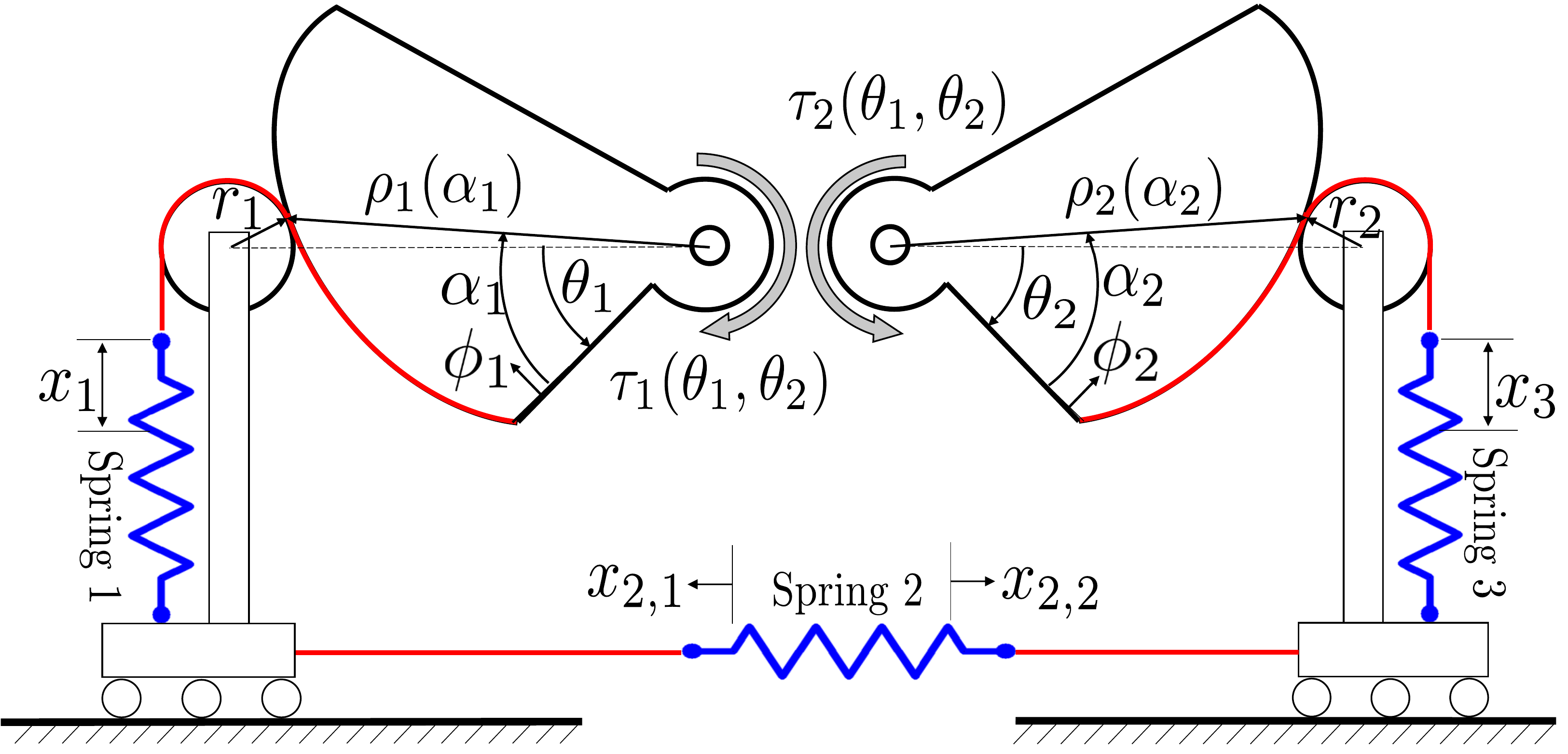}
    \caption{Cam design concept for a two DOF system}
    \label{fig:cam_2dof}
\end{figure}

A two DOF design variant is shown in Fig. \ref{fig:cam_2dof}. This system is designed so that the torque on the cams are functions of both cam angles (i.e. $\tau_1(\theta_1,\theta_2)$ and $\tau_1(\theta_1,\theta_2)$). A planar RR manipulator arm is a potential application of this design variant. This system contains two cams with very similar structure to the one DOF cam system shown in Fig. \ref{fig:cam_1dof}. However, instead of being connected to ground, spring 2 connects the prismatic platforms for both cams. This means that changing $\theta_1$ affects the torque on cam 2 and vice versa. In this section, we again assume infinite friction between the cam and wire and assume that the springs are linear functions of displacement.
\par Because the angles $\alpha_1$ and $\gamma_1$ are independent of $\theta_2$ and $\alpha_2$ and $\gamma_2$ are independent of $\theta_1$, the equations in Section \ref{sec:alpha_gamma} can be directly applied to solve for the tangency angles of each cam.
%
\par Next, we will derive the spring displacements $x_1$, $x_2$, and $x_3$. Because the displacement of springs 1 and 3 do not depend on the angle of the cam they are not wrapped around, Eq. \eqref{eq:x1} can be used to solve for the spring displacements $x_1$ and $x_3$ as shown in Fig. \ref{fig:cam_2dof}. Because the displacement of spring 2 depends on both $\theta_1$ and $\theta_2$, we must apply Eq. \eqref{eq:x2} to solve for the displacement of spring 2 while taking into account the individual contributions of each cam. To designate the contribution of cam $i$  to the extension of spring $2$ we use the notation $x_{2,i}$, $i=1,2$. Referring to Figure~\ref{fig:cam_2dof}, we can write the extension of the second spring
\begin{equation}\label{eq:x2_2dof}
   x_{2} =  x_{2_0}+x_{2,1} + x_{2,2}
\end{equation}
where $x_{2_0}$ is the initial spring extension and the individual cam contributions to the spring extension are given by:
\begin{equation}\label{eq:x2_2dof}
\begin{aligned}
   x_{2,i} =& \rho_i(\alpha_i)\cos(\alpha_i-\theta_i)-r_i\cos(\gamma_i) \\
   &-(\rho_i(\alpha_{i_0})\cos(\alpha_{i_0})-r_i\cos(\gamma_{i_0})), \quad i = 1,2
\end{aligned}
\end{equation}
Where the subscript $i$ indicates the cam number. Having expressed the spring extensions due to cam rotations, we next compute the resulting balancing torques. In the beginning of Section \ref{sec:one_dof_torque} we consider a model using a practice prevalent in the literature, which assumes infinite friction between the cam and wire. In Section \ref{sec:finite_friction} we relax the assumption of infinite friction and present a model that accounts for a finite friction coupling between the cam and wire.

\section{One DOF Wire-Wrapped Cam Torque}\label{sec:one_dof_torque}
Substituting the solution of Eq.~\eqref{eq:alpha_gamma} for $\alpha$ and $\gamma$ in Eqs.~\eqref{eq:x1} and~\eqref{eq:x2}  provides the spring extensions $x_1$ and $x_2$. Assuming infinite friction coupling and referring to Fig.~\ref{fig:cam_1dof},  the force $\mb{f}_1$ applied by the wire rope on the cam is concentrated at the tangency point $\mb{p}$ and pointing tangential to the cam:
\begin{equation}\label{eq:f1}
  \mb{f}_1 = k_1x_1\hat{\mb{t}}
\end{equation}
In this equation, $k_1$ is the stiffness of spring 1 and $\hat{\mb{t}} = \hat{\mb{t}}(\alpha) = -\hat{\mb{t}}(\gamma)$ is a unit vector tangent to the idler/cam contact given by Eq. \eqref{eq:t}. The moment of spring 1 on the cam is the cross-product of $\mb{r}_{p/o}$ and $\mb{f}_1$. Similarly, the force applied by spring 2 on the cam is:
\begin{equation}\label{eq:f2}
  \mb{f}_2 = k_2x_2\hat{\mb{n}}
\end{equation}
where $k_2$ is the stiffness of spring 2 and $\hat{\mb{n}} = \frac{d}{d\alpha}\hat{\mb{t}}(\alpha) = -\frac{d}{d\gamma}\hat{\mb{t}}(\gamma)$ is the unit vector normal to the idler/cam contact. The moment of spring 1 on the cam is the cross-product of $\mb{r}_{p/o}$ and $\mb{f}_2$.
\par The scalar moment acting on the cam is given by the spring force moments about point $\mb{o}$, therefore:
\begin{equation}\label{eq:tau}
  \tau_1 = (\underbrace{\mb{r}_{p/o}\times\mb{f}_1}_{\bs{\tau}_{1,1}}+\underbrace{\mb{r}_{p/o}\times\mb{f}_2}_{\bs{\tau}_{1,2}})\T\,\uvec{z}
\end{equation}
where the single subscript in $\tau_1$ designates the cam number. It is included to maintain consistency with the notation for the two DOF case. Also, in $\bs{\tau}_{1,1}$ and $\bs{\tau}_{1,2}$ the first subscript refers to the cam number and the second subscript refers to the spring number.
\subsection{Wire-Wrapped Cam Torque with Finite Friction Coupling}\label{sec:finite_friction}
In this section, we explore the effect of finite capstan/wire-rope friction on the resulting torque on the cam. We model the forces at a quasi-static equilibrium where we assume $\theta$ is fixed.
\subsubsection{Wire Model}
We will assume that the wire is inextensible and can only support tension forces (i.e. it cannot support transverse shear forces and bending moments). The tendon curve is assumed to match the planar cam shape parameterized by angle $\phi$  as shown in Fig.~\ref{fig:cam_1dof}b.
\begin{figure}[h]
    \centering
    \includegraphics[width=0.7\columnwidth]{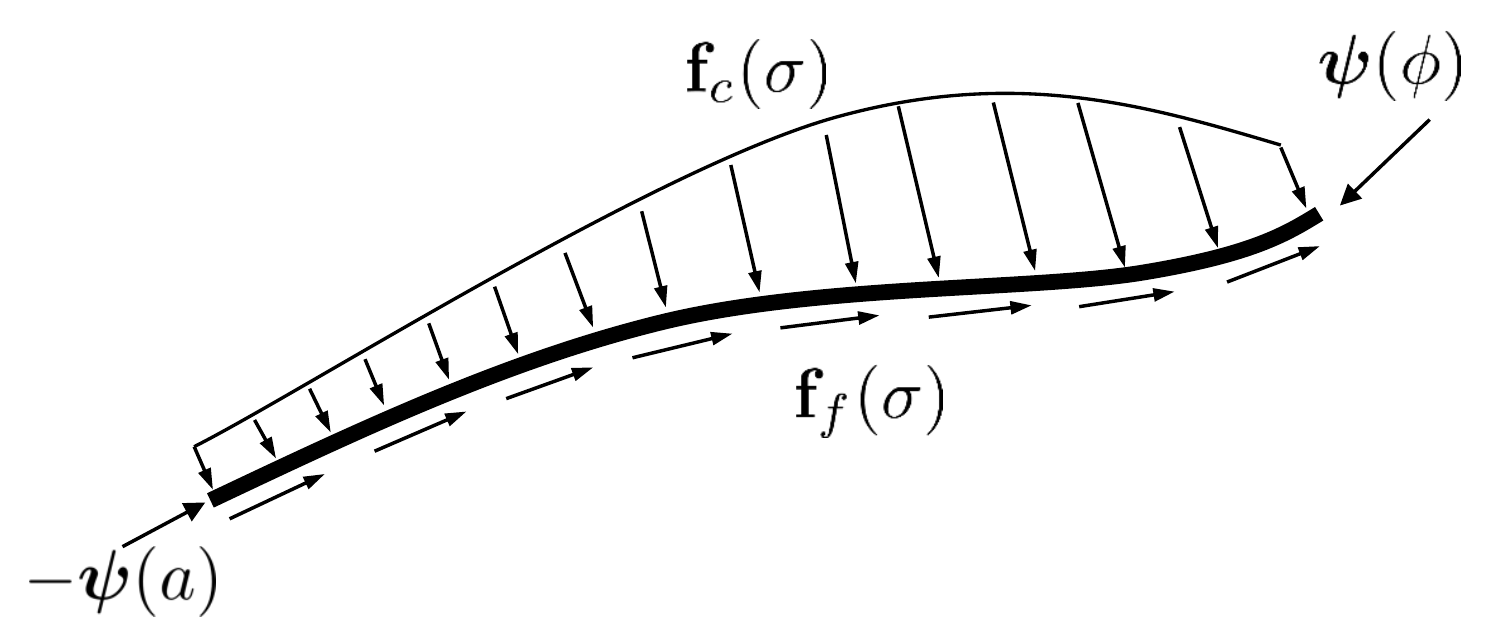}
    \caption{Free body diagram of the wire with distributed loads and internal forces.}
    \label{fig:wire_fbd}
\end{figure}
\par  Following \cite{antman2004nonlinear}, we define the wire \emph{internal force} $\bs{\psi}(\phi) \in \realfield{3}$ as the force applied by the material at $\phi + \delta\phi$ on the material at $\phi$. To be bent into the cam curve, the wire must have an external force or moment distribution applied to it. Referring to Fig. \ref{fig:wire_fbd}, we use $\mb{f}_c$ to designate the distributed normal/contact force between the wire and the cam and $\mb{f}_f$ as the distributed friction force between the wire and the cam.
%
The force balance for a section $[a,\phi]$ is given by:
\begin{equation}
\bs{\psi}(\phi) -\bs{\psi}(a) + \int_{a}^{\phi} \mb{f}(\sigma)\text{d}\sigma = 0
\end{equation}
where $\mb{f}$ is the resultant of the contact and friction force on the wire (per unit of wrapping angle $\phi$):
\begin{equation}
\mb{f} = \mb{f}_f+\mb{f}_c
\end{equation}
 We then take a derivative of this with respect to $\phi$:
\begin{equation}\label{eq:tension_derivative}
\bs{\psi}'(\phi) + \mb{f}(\phi) = 0
\end{equation}
where $(')$ denotes a derivative with respect to $\phi$.
\par  Assuming the tendon is always in contact with the cam, and since the internal force is always tangent to the wire path (since the wire cannot support transverse shear and moment loads), we can write the internal force in the wire:
\begin{equation}  \label{eq:wire_internal_force}
\bs{\psi}(\phi) = \eta(\phi)\frac{\mb{r}'(\phi)}{\| \mb{r}'(\phi) \|}
\end{equation}
where $\frac{\mb{r}'(\phi)}{\| \mb{r}'(\phi) \|}$ designates the local tangent, which can be found using a similar derivation of $\mb{t}(\alpha)$ in Eq.~\eqref{eq:tangent_partials}, but for \mbox{$\alpha-\theta=\phi$}. Also, $\eta(\phi)>0$ designates the wire tension and $\mb{r}(\phi)$  denotes the polar vector in a cam-attached frame.
\begin{equation}\label{eq:r}
  \mb{r}(\phi) = \rho(\phi)\left[
             \cos(\phi) ,~
             \sin(\phi),~
             0\right]\T
\end{equation}
\subsubsection{Capstan Equation}
\begin{figure}[h]
    \centering
    \includegraphics[width=0.7\columnwidth]{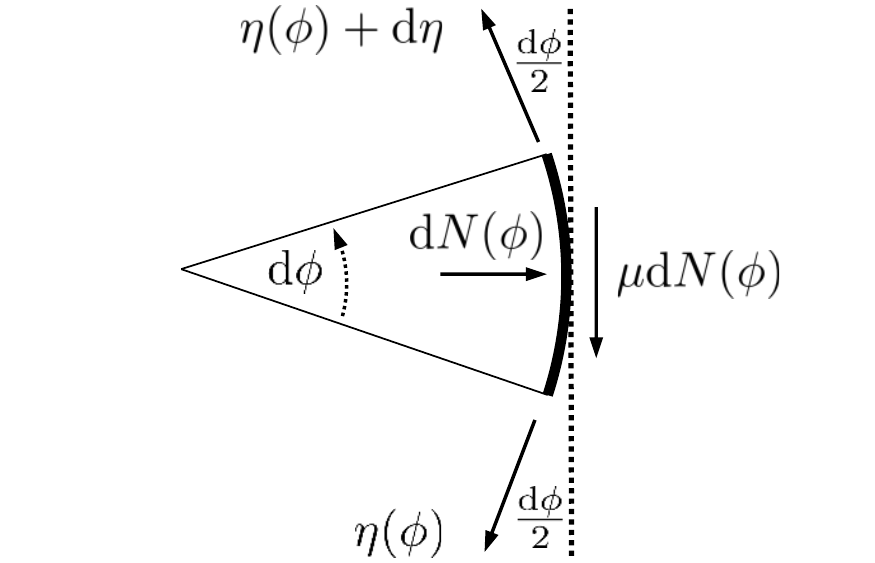}
    \caption{Differential forces due to friction between the cam and the wire.}
    \label{fig:capstan_fbd}
\end{figure}

We next derive the tension distribution $\eta(\phi)$ following the approach in \cite{meriam1987mechanics}.
Referring to the wire rope segment shown in Fig. \ref{fig:capstan_fbd}, we define the distributed normal force $dN$ over a wire segment given by $d \phi$ and the tension force magnitudes given by $\eta(\phi)$ and $\eta(\phi)+d\eta$. Writing the force balance on this infinitesimal wire segment yields:
\begin{subequations}
\begin{align}
&\text{d}N = \left(\eta + \text{d}\eta\right)\sin{}\frac{\text{d}\phi}{2} + \eta\sin{}\frac{\text{d}\phi}{2}\\
&\left(\eta + \text{d}\eta \right)\cos{}\frac{\text{d}\phi}{2} = \eta\cos{}\frac{\text{d}\phi}{2} + \mu\text{d}N
\end{align}
\end{subequations}
\par \noindent Applying small angle approximations $\sin\phi \approx \phi$, $\cos\phi \approx 1$ and neglecting the product of two small differentials, we get:
\begin{subequations}
\begin{align}
\text{d}N &= \left(\eta + \text{d}\eta\right)\frac{\text{d}\phi}{2} + \eta\frac{\text{d}\phi}{2} \approx \eta\text{d}\phi \label{eq:dN}\\
\mu\text{d}N &= \text{d}\eta  \label{eq:deta}
\end{align}
\end{subequations}
Substituting Eq. \eqref{eq:dN} into Eq. \eqref{eq:deta}, separating variables, and integrating from the initial angular coordinate $\phi$ up to the roller/cam tangency point (designated by angular coordinate $\alpha$) results in:
\begin{equation}
 \int_{\phi}^{\alpha}\mu \; \text{d}\phi = \int_{\eta(\phi)}^{\eta(\alpha)} \frac{1}{\eta} \; \text{d}\eta
 \end{equation}
 Writing the above integrals and taking the exponent, gives:
 \begin{equation}
 e^{\mu(\alpha - \phi)} = \frac{\eta(\alpha)}{\eta(\phi)}
\end{equation}
Solving for the wire tension $\eta(\phi)$ at any location $\phi$ gives:
\begin{equation}\label{eq:eta_of_phi}
 \eta(\phi) = \eta(\alpha)e^{\mu\left( \phi- \alpha \right)} \quad \phi \in [0,\alpha]
\end{equation}
Next, we will use this result to compute the distributed force in the wire and the resulting torque on the cam.
\subsubsection{Distributed Wire Force and Cam Torque}
\noindent The wire force vector was given by Eq. \eqref{eq:wire_internal_force}. Substituting Eq. \eqref{eq:eta_of_phi} in Eq. \eqref{eq:wire_internal_force} gives:
\begin{equation}
\bs{\psi}(\phi) = \eta(\alpha)e^{\mu\left(\phi - \alpha  \right)}\frac{\mb{r}'(\phi)}{\| \mb{r}'(\phi) \|}
\end{equation}
Differentiating the above equation with respect to $\phi$ gives:
\begin{equation}\label{eq:f_phi}
\bs{\psi}'(\phi) = \frac{\eta(\alpha)e^{\mu\left(\phi - \alpha  \right)}}{\| \mb{r}{'} \| }     \left(  \mu\mb{r}{'} + \mb{r}{'}{'} - \mb{r}{'}(\mb{r}{'}\T\mb{r}{'})^{3}(\mb{r}{'}{'}\T\mb{r}{'}) \right)\\
\end{equation}
Using the above result in Eq. \eqref{eq:tension_derivative}, we obtain the resultant distributed force on the wire:
\begin{equation}
\mb{f}(\phi)=-\frac{\eta(\alpha)e^{\mu\left(\phi - \alpha  \right)}}{\| \mb{r}{'} \| }     \left(  \mu\mb{r}{'} + \mb{r}{'}{'} - \mb{r}{'}(\mb{r}{'}\T\mb{r}{'})^{3}(\mb{r}{'}{'}\T\mb{r}{'}) \right)
\end{equation}
\par The torque applied by the wire on the cam is given by the moment arm of the force $\bs{\psi}$ at the wire-anchor-point $\phi=0$ and the moment of the distributed force $\mb{f}(\phi)$:
\begin{equation}\label{eq:tau1_fric}
\bs{\tau}_{1,1} =\mb{r}(0) \times  \bs{\psi}(0)+ \int_{0}^{\alpha}  \mb{r}(\sigma) \times -\mb{f}(\sigma) \; \text{d}\sigma
\end{equation}
%
This is an alternate formulation for $\bs{\tau}_{1,1}$ given in Eq. \eqref{eq:tau} that now removes the assumption of infinite friction.

\section{Two DOF Wire-Wrapped Cams}\label{sec:2dof_torque}
Regardless of the type of friction coupling between the cam and the wire-rope, the results for the two DOF case are simply a superposition of the reactions applied by each spring on each cam. Referring to Figure~\ref{fig:cam_2dof}, we note that cam 1 is subject to torque contributions from spring 1 and spring 2. Similarly, cam 2 is subject to torque contribution due to reactions from spring 2 and spring 3. Adopting the notation  $\bs{\tau}_{i,j}$ to designate the torque contribution of spring $j$ on cam $i$, we can write:
\begin{equation}\label{eq:torque_2dof}
  \tau_i = \left(\bs{\tau}_{i,i} + \bs{\tau}_{i,i+1}\right)\T\hat{\mb{z}} \quad i= 1,2
\end{equation}
For the two DOF cam system shown in Fig. \ref{fig:cam_2dof}, the torque on cam 1 and cam 2, $\tau_1$ and $\tau_2$, can be found by applying the equations from Section \ref{sec:one_dof_torque} using the spring displacements from Section \ref{sec:2dof_deflection}.
$\bs{\tau}_{2,3}$ should be calculated using Eqs.~\eqref{eq:tau}  or \eqref{eq:tau1_fric}.
\par We will next present considerations for cam convexity, which will feed into the cam optimization process.   
\section{Modal Cam Shape and Condition for Convexity}\label{sec:modal}
In this work, we represent the profile of the cam profile using a polar representation:
\begin{equation}\label{eq:xy_polar}
  x(\phi) = \rho(\phi)\cos(\phi) \qquad y(\phi) = \rho(\phi)\sin(\phi)
\end{equation}
where $\rho$ and $\phi$ are the radial and polar coordinates. The radial coordinate $\rho(\phi)$ can be expressed via a modal representation:
\begin{equation}\label{eq:rho}
  \rho(\phi) = \bs{\beta}\T
  \begin{bmatrix}
    1, & \phi, & \phi^2, & \ldots, & \phi^{n-1}, & \phi^n \end{bmatrix}\T
\end{equation}
\noindent Where $\bs{\beta}\in\realfield{n+1}$ is a vector of polynomial coefficients.
\par In making the above choice of using a polynomial basis, we implicitly limit ourselves to cam representations with orders below $n = 7$ since polynomial bases are known to suffer from poor numerical conditioning for large powers ($n>7$) \cite{Angeles2012_cam_optimization}.
\par In order for the wire to maintain contact with the cam, the cam should be convex at all points along its surface. As shown in \cite{Bajaj1991}, a curve $f\colon(x(\phi),y(\phi))\rightarrow\realfield{2}$ is convex if:
\begin{equation}\label{eq:convex}
  x'(\phi)y''(\phi)-y'(\phi)x''(\phi) > 0, \quad \forall \phi
\end{equation}
where $(\cdot)'$ and $(\cdot)''$ designate first and second derivatives with respect to $\phi$.
\noindent Using Eq. \eqref{eq:xy_polar} in  Eq. \eqref{eq:convex} yields the  constraint for ensuring cam convexity:
\begin{equation}\label{eq:convex_polar}
  \rho(\phi)^2+2\rho'(\phi)^2-\rho(\phi)\rho''(\phi) > 0
\end{equation}
\section{Sensitivity to Changes in Spring Constant}\label{sec:sens}
If a cam design has one spring with constant $k$, the cam torque deviation $\Delta\tau$ for a fixed cam angle that ensues due to spring constant deviation $\Delta k$ can be found using a first-order approximation \cite{belegundu1992}:
\begin{equation}\label{eq:sens}
  \Delta\tau = \frac{\partial\tau}{\partial k}\Delta k
\end{equation}
\noindent For the one DOF cam, to minimize the effect of deviations in spring constant, the design parameters (the cam modal coefficients $\bs{\beta}$ and the spring pre-extensions $\mb{x}_0$) are selected to minimize the following weighted sum:
\begin{equation}\label{eq:sens_opt1}
\underset{\bs{\beta}, \mb{x}_0}{\min}\, \left(\sum_{i=1}^2 w_{1,i} \int_{\theta_{min}}^{\theta_{max}}\Big|\frac{\partial\tau_1}{\partial k_i}\Big|d\theta\right)
\end{equation}
where  $\tau_1$, is given by Eq. \eqref{eq:tau}. The scalar weights $w_{1,1}$ and $w_{1,2}$ represent the relative importance of minimizing the effect of unmodeled deviations in the stiffness of spring 1 and 2, respectively. For example, larger weights can be set for springs with larger uncertainty in their spring constant.
\par For the infinite friction case, we note that $\frac{\partial \bs\tau_{1,1}}{\partial k_2}= \frac{\partial \bs\tau_{1,2}}{\partial k_1} = \mb{0}$ and use the commutativity of the dot product in Eq. \eqref{eq:tau}, to obtain:
\begin{equation}\label{eq:dtau_dk}
  \frac{\partial \tau_1}{\partial k_1} = \hat{\mb{z}}\T\left(\mb{r}_{p/o}\times x_1\hat{\mb{t}}\right),\quad
  \frac{\partial \tau_1}{\partial k_2} = \hat{\mb{z}}\T\left(\mb{r}_{p/o}\times x_2\hat{\mb{n}}\right)
\end{equation}
\par For the finite friction case, the partial of $\tau_1$ with respect to $k_2$ remains the same as in Eq. \eqref{eq:dtau_dk}. However, the partial of $\tau_1$ with respect to $k_1$ needs to be updated with the equations in Section \ref{sec:finite_friction}. Given that the wire tension at the contact point is $\eta(\alpha) = k_1x_1$, this can be found by taking the partial of Eq. \eqref{eq:tau1_fric}:
\begin{equation}\label{eq:tau1_sens_1dof}
\begin{aligned}
  &\frac{\partial}{\partial k_1}\tau_{1} = \uvec{z}\T \Big(\mb{r}(0) \times  x_1e^{-\mu\alpha}\frac{\mb{r}'(0)}{\| \mb{r}'(0) \|}\\
  &+ \int_{0}^{\alpha}  \mb{r}(\sigma) \times \frac{x_1e^{\mu\left(\sigma - \alpha  \right)}}{\| \mb{r}{'} \| }     \left(  \mu\mb{r}{'} + \mb{r}{'}{'} - \mb{r}{'}(\mb{r}{'}\T\mb{r}{'})^{3}(\mb{r}{'}{'}\T\mb{r}{'}) \right) \; \text{d}\sigma\Big)
\end{aligned}
\end{equation}
\par The two DOF design in Fig.~\ref{fig:cam_2dof} has three springs and two cams, the objective function is therefore modified as:
\begin{equation}\label{eq:sens_opt2}
\begin{aligned}
 \underset{\bs{\beta}_1, \bs{\beta}_2, \mb{x}_0}{\min}\, \Bigg( &\sum_{i=1}^{3}w_{1,i}\int_{\theta_{2,min}}^{\theta_{2,max}}\int_{\theta_{1,min}}^{\theta_{1,max}}\Big|\frac{\partial\tau_1}{\partial k_i}\Big|d\theta_1 d\theta_2 \\ +&\sum_{i=1}^{3}w_{2,i}\int_{\theta_{2,min}}^{\theta_{2,max}}\int_{\theta_{1,min}}^{\theta_{1,max}}\Big|\frac{\partial\tau_2}{\partial k_i}\Big|d\theta_1 d\theta_2
 \Bigg)
\end{aligned}
\end{equation}
\noindent In this equation, $\tau_i$, $i=1,2$, are given by Eq. \eqref{eq:torque_2dof}  and $w_{1,i}$ $w_{2,i}$  are scalar weights representing the relative importance of minimizing the effect of uncertainty in spring $i$ on torques for cams 1 and 2. 
\section{Optimal Wire-Cam Mechanism Design}\label{sec:opt}
\subsection{One DOF Cam System}\label{subsec:one_dof_optimal_cam}
Now that we can calculate the torque on the cam from Section \ref{sec:one_dof_torque} and the sensitivity of the torque to unmodeled changes in the spring constant from Section \ref{sec:sens}, we can formulate the optimization problem used to design the cam system. The optimization variables used for this problem are the cam polynomial coefficients $\bs{\beta}$ and the spring pre-extensions $\mb{x}_0 = \begin{bmatrix} x_{1_0} & x_{2_0} \end{bmatrix}$. The following parameters are taken to be constants: the idler radius $r$, the spring constants $k_i,~i=1,2$, the maximum spring deflections $x_{i,max}$, and the wire-cam coefficient of friction $\mu$. The desired cam torque $\tau_d(\theta)$, which is the torque needed to statically balance the joint, is also taken as an input. The condition for convexity from Section \ref{sec:modal} and practical limits on the cam geometry and spring deflections are taken as constraints:
\begin{equation}\label{eq:opt}
\begin{aligned}
\min_{\bs{\beta}, \mb{x}_{0}} \quad & w_1\int_{\theta_{min}}^{\theta_{max}}
\left(\tau_{d,1}(\theta)-\tau_1(\theta)\right)^2d\theta \\
 & + \sum_{i=1}^{2}w_{i+1}\int_{\theta_{min}}^{\theta_{max}}\left|\frac{\partial\tau_1(\theta)}{\partial k_i}\right|d\theta \\
\textrm{s.t.} \quad & \rho^2+2\rho'^2-\rho\rho''> 0 \\
  & \rho_{min} < \rho < \rho_{max}   \\
  & 0 \leq x_i < x_{i,max},~i=1,2 \\
  & \mb{0} \leq \mb{x}_{0}  \\
\end{aligned}
\end{equation}
\noindent Where $w_1$, $w_2$, and $w_3$  are scalar weights which represent the relative importance of the terms.
\subsection{Two DOF Cam System}\label{subsec:two_dof_optimal_cam}
For the two DOF system, the desired torque on the cams may be functions of both cam angles: $\tau_{d,1}(\theta_1,\theta_2)$ and $\tau_{d,2}(\theta_1,\theta_2)$. The cam 1 polynomial coefficients $\bs{\beta}_1$, cam 2 polynomial coefficients $\bs{\beta}_2$ and spring pre-extensions $\mb{x}_0 = \begin{bmatrix} x_{1_0} & x_{2_0} & x_{3_0} \end{bmatrix}$ are the optimization variables. We also assume that the desired cam torques $\tau_{d,1}(\theta_1,\theta_2)$ and $\tau_{d,2}(\theta_1,\theta_2)$ are specified by the design task (i.e., the static balancing torques). In the equation below, the arguments from $\tau_i(\theta_1, \theta_2),~i=1,2$ and $\tau_{d,i}(\theta_1, \theta_2)~i=1,2$ are dropped for notational brevity.
\begin{equation}\label{eq:opt_2dof}
\begin{split}
\min_{\bs{\beta}_1, \bs{\beta}_2, \mb{x}_{0}} \,\, & \int_{\theta_{min,2}}^{\theta_{max,2}}\int_{\theta_{min,1}}^{\theta_{max,1}}
\sum_{j=1}^{2}w_j\left(\tau_{d,j}-\tau_j\right)^2 d\theta_1d\theta_2 \\
&+\sum_{i=1}^{3}w_{i+2}\int_{\theta_{min,2}}^{\theta_{max,2}}\int_{\theta_{min,1}}^{\theta_{max,1}}\left|\frac{\partial\tau_1}{\partial k_i}\right|d\theta_1d\theta_2 \\
&+\sum_{i=1}^{3}w_{i+5}\int_{\theta_{min,2}}^{\theta_{max,2}}\int_{\theta_{min,1}}^{\theta_{max,1}}\left|\frac{\partial\tau_2}{\partial k_i}\right|d\theta_1d\theta_2 \\
\textrm{s.t.} \quad & \rho_i^2+2\rho_i'^2-\rho_i\rho_i''> 0,~i=1,2 \\
  & \rho_{min} < \rho_i < \rho_{i,max},~i=1,2   \\
  & 0\leq x_i < x_{i,max},~i=1,2,3 \\
  & \mb{0} \leq \mb{x}_{0}  \\
\end{split}
\end{equation}
\noindent \remind{R1-12}{Where $w_1$ through $w_8$ are scalar weights which represent the relative importance of the terms. For example, a spring with a loose manufacturer-specified stiffness tolerance would require a higher weight to reduce the effect of this variability on the cam torque error.}   
\section{Wire-Cam Coefficient of Friction Characterization}\label{sec:fric_char}
\begin{figure}[htbp]
    \centering
    \includegraphics[width=0.8\columnwidth]{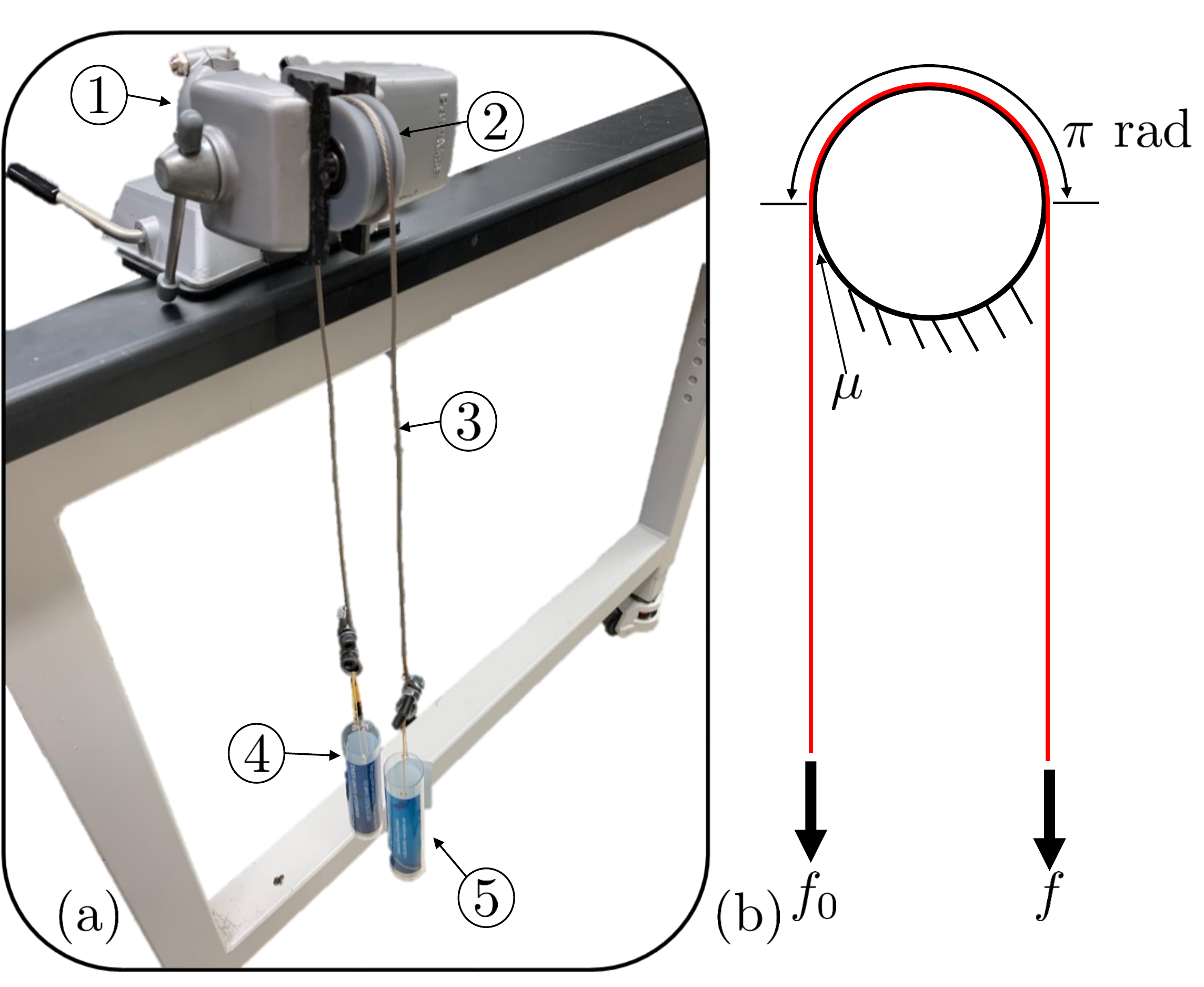}
    \caption{(a) Friction characterization experimental setup: \protect\circled{1} benchtop vise, \protect\circled{2} cylindrical portion of cam material, \protect\circled{3} wire-rope, \protect\circled{4} basket for holding mass that creates $f_0$, \protect\circled{5} basket for holding mass that creates $f$. (b) Schematic of experimental setup showing terms used in Eq.~\eqref{eq:mu}}
    \label{fig:fric_setup}
\end{figure}
Because the static coefficient of friction is used as an input into Eq.~\eqref{eq:opt_2dof}, we must characterize the static coefficient of friction between the cam and the wire before we can use Eq.~\eqref{eq:opt_2dof} to solve for the cam shape. In this work, we 3D printed the cams using Formlabs\texttrademark~Grey Pro resin and chose the wire to be a stainless steel, 7$\times49$ construction, 0.08 inch diameter wire-rope. To test the static coefficient of friction between these materials, we printed a cylindrical portion of the cam material (\circled{2} in Fig. \ref{fig:fric_setup}) and draped the wire over the cylinder such that it was wrapped $180^\circ$ along the cylinder's circumference as shown in Fig. \ref{fig:fric_setup}. A small force $f_0$ was added to one end of the wire and then a force was added to the other end of the wire until static friction was broken and the wire slipped. The slippage force $f$ was recorded and the coefficient of friction was calculated using the capstan equation:
\begin{equation}\label{eq:mu}
  \mu = \frac{1}{\pi}\operatorname{ln}\left(\frac{f}{f_0}\right)
\end{equation}
This experiment was repeated 10 times. The mean coefficient of friction was found to be 0.3273 with a standard deviation of 0.0274. 
\section{Simulation Case Studies}\label{sec:sim}
\subsection{Comparison to Closed-Form Solution}\label{sec:comparison}
\begin{figure}[h]
  \centering
  \includegraphics[width=\columnwidth]{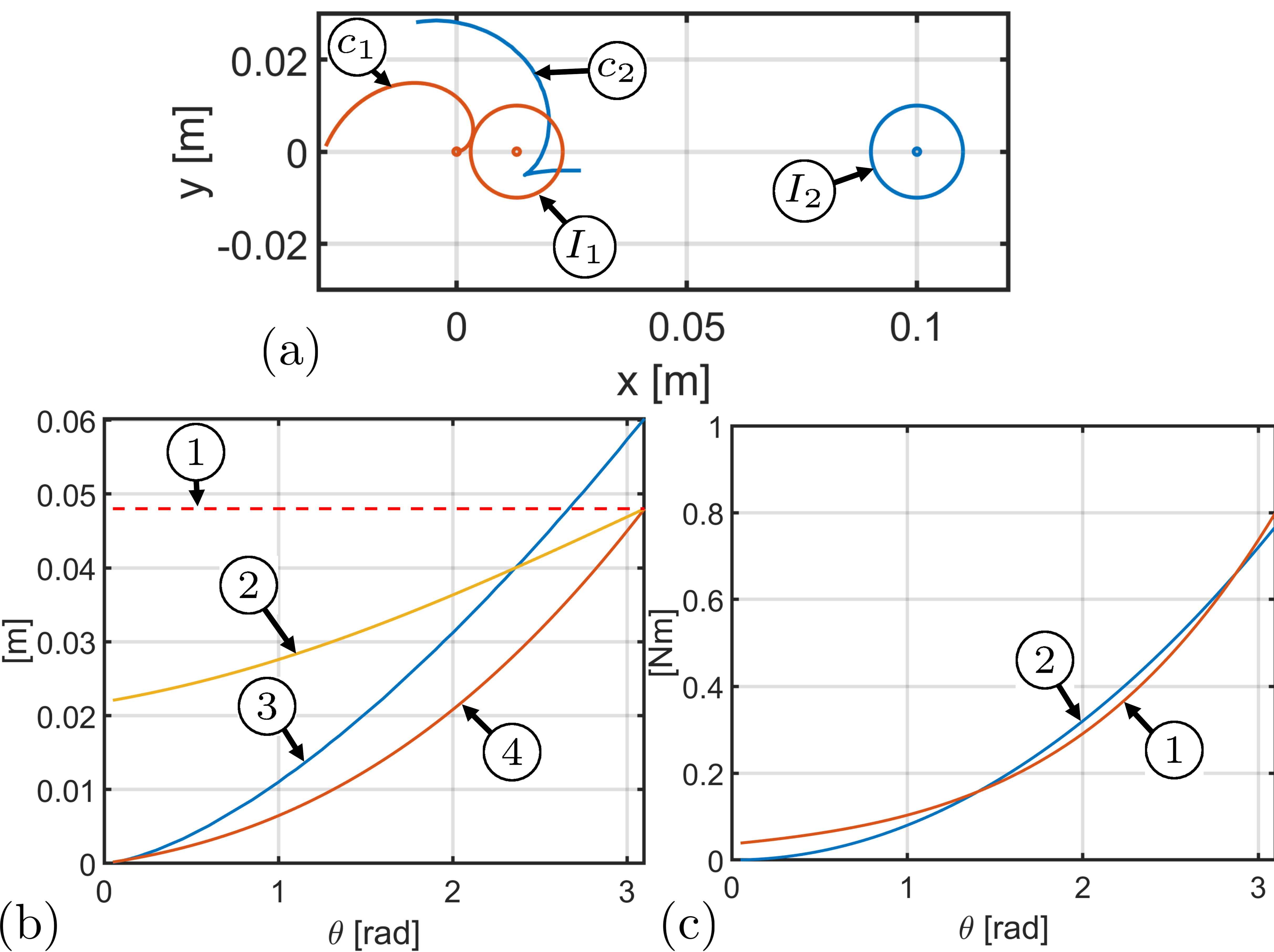}
  \caption{(a) Cam shapes and idlers for our method ($c_1$ and $I_1$) and \cite{Kilic2012} ($c_2$ and $I_2$) when $\theta = 0$. (b) spring deflections: \protect\circled{1} Maximum allowable spring deflection, \protect\circled{2} spring 2 deflection $x_2$ for our method, \protect\circled{3} spring deflection in \cite{Kilic2012}, and \protect\circled{4} spring 1 deflection $x_1$ for our method and (c) desired vs. actual torque cam torque for our method: \protect\circled{1} our method, \protect\circled{2} desired torque.}\label{fig:kilic}
\end{figure}
\par \corrlab{R1-3}{To demonstrate the benefit of our approach relative to the closed-form solution, we offer here an example of a single DOF cam design for satisfying the desired torque function $\tau_d = 0.08\theta^2$. Figure \ref{fig:kilic} shows a comparison of our method to the method of \cite{Kilic2012} with an assumed idler horizontal location of 0.1 m, McMaster-Carr spring (part number 5108N486), and idler radius of 0.01 m. For the method of \cite{Kilic2012}, the spring was assumed to have no pre-extension.
\par The initial conditions of our optimization-based approach were $\mb{x}_0 = [10,~10]\T$ mm and $\bs{\beta} = [0.1,~0.1,~0.1,~0.1]\T$. The scalar weights were $w_1 = 10,~w_2 = w_3 = 0$. No size limit was placed on the cam.
\par Obviously, the closed-form cam profile \circled{$c_2$} shown in Fig. \ref{fig:kilic}(a) is not convex and would pose a problem for design implementation. Our method, on the other hand remains convex (cam \circled{$c_1$}). Additionally, as can be seen in Fig. \ref{fig:kilic}(b), the method in \cite{Kilic2012} does not respect the extension limits of the spring. Our method, on the other hand, does not violate the extension limits. This more realistic design comes at the cost of 27.87 Nmm of root-mean-square error (RMSE) between $\tau$ and $\tau_d$.}
%
\subsection{Desired Torque Functions}
\begin{figure}
  \centering
  \includegraphics[width=0.60\columnwidth]{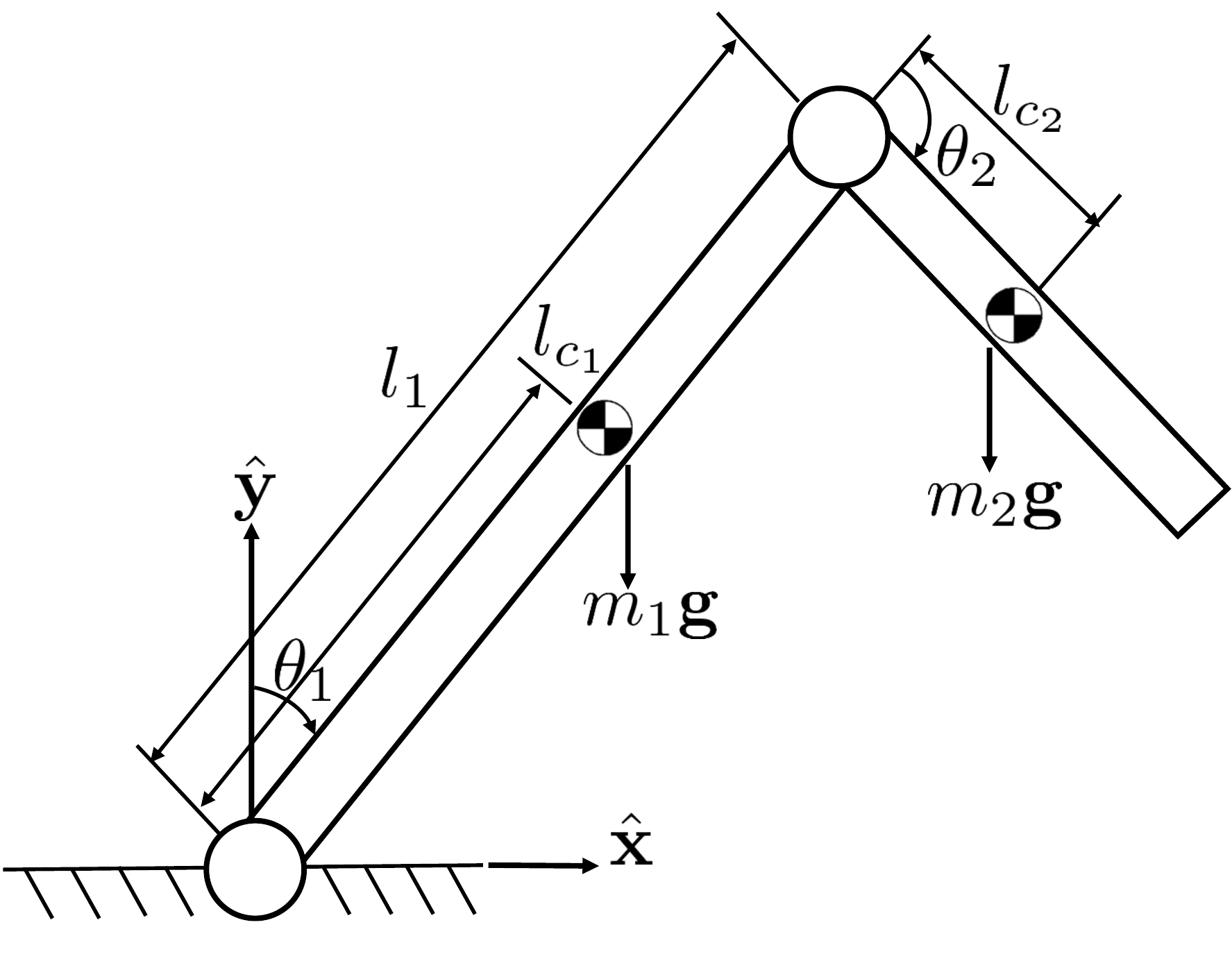}
  \caption{RR manipulator arm used to generate the desired torque functions.}\label{fig:RR}
\end{figure}
To demonstrate our method, we will solve the optimization problem of Eq. \eqref{eq:opt_2dof} for the following desired torque functions:
\begin{equation}\label{eq:taud}
\begin{aligned}
\tau_{d,1} &= m_1gl_{c_1}\sin(\theta_1) + m_2g\left[l_1\sin(\theta_1)+l_{c_2}\sin(\theta_1+\theta_2)\right] \\
\tau_{d,2} &= m_2gl_{c_2}\sin(\theta_1+\theta_2)
\end{aligned}
\end{equation}
These equations represent the torque required to balance the gravitational load on the joints of the RR manipulator arm shown in Fig. \ref{fig:RR}. In this equation, $m_1$ is the mass of link 1; $l_{c_1}$ is the location of the center of mass of link 1; $l_1$ is the length of link 1; $m_2$ is the mass of link 2; and $l_{c_2}$ is the center of mass of link 2. The values of the aforementioned link masses and lengths are summarized in Table \ref{tab:tau_des}.
\begin{table}[htbp]
\renewcommand{\arraystretch}{1.0}
\fontsize{8pt}{10pt}\selectfont  
\caption{Desired torque function constants}\label{tab:tau_des}
\small
\begin{center}
\begin{tabular}{ |c|c|c|c|c|c|}
 \hline
 $m_1$ [kg] & $m_2$ [kg] & $l_{c_1}$ [m] & $l_1$ [m] & $l_{c_2}$ [m] & $\theta$\\\hline
 0.5  &  0.5 & 0.25 &  0.5 & 0.25 & $[0^\circ,~ 90^\circ]$  \\\hline
\end{tabular}
\end{center}
\end{table}
\par We will solve this optimization problem twice: once without trying to minimize the sensitivity of the cam torque to unmodeled changes in the spring constant and once allowing the optimization routine to minimize the aforementioned sensitivity. In the next section, we will discuss the constants and initial conditions used in both simulations.
\subsection{Simulation Constants and Initial Conditions}\label{subsec:sim_IC}
\par While in some cases it may be desirable to let the optimization routine select the spring geometries, the springs in this case study were pre-selected and used as constants. The springs were chosen from McMaster-Carr stock springs. Spring one was selected to be part number 5667N212; spring two was selected to be part number 7749N634; and spring three  was selected to be part number 1942N653. The rate and maximum extensions of these springs are shown in Table \ref{tab:sim_consts} along with the idler radii, coefficient of friction between the cam and the wire, minimum and maximum cam radii, and the idler vertical offsets.
\begin{table*}[htbp]
\renewcommand{\arraystretch}{1.0}
\fontsize{8pt}{10pt}\selectfont  
\caption{Simulation constants}\label{tab:sim_consts}
\begin{center}
\begin{tabular}{|c|c|c|c|c|c|c|c|c|}
 \hline
 $k_1,~k_2,k_3$ [N/mm] & $x_{1,max},~x_{2,max},~x_{3,max}$ [mm] & $r_1,~r_2$ [mm] & $\mu$ & $\rho_{min},~\rho_{max}$ [mm], & $a_{0_1},~ a_{0_2}$ [mm]\\\hline
 1.10, 7.35, 0.58 &  57.66, 32.00, 105.00 & 20, 20 & 0.3273 & 25, 500 & 15, 15 \\\hline
\end{tabular}
\end{center}
\end{table*}
The optimization problems were solved using the active-set algorithm implemented in Matlab\texttrademark~2021b's \textit{fmincon} function. \corrlab{R1-1-3}{The computer used for these simulations is running Windows 10 on an Intel i7-7700 3.6 GHz processor and has 16 Gb of ram. }The initial conditions assumed for this simulation were 10 mm of pre-extension for all springs and $\bs{\beta}_1 = \bs{\beta}_2 = [0.001,0.001,0.001,0.001]\T$.
\subsection{Two DOF System Without Minimizing Sensitivity}
\begin{figure*}[htbp]
    \centering
    \includegraphics[width=0.9\textwidth]{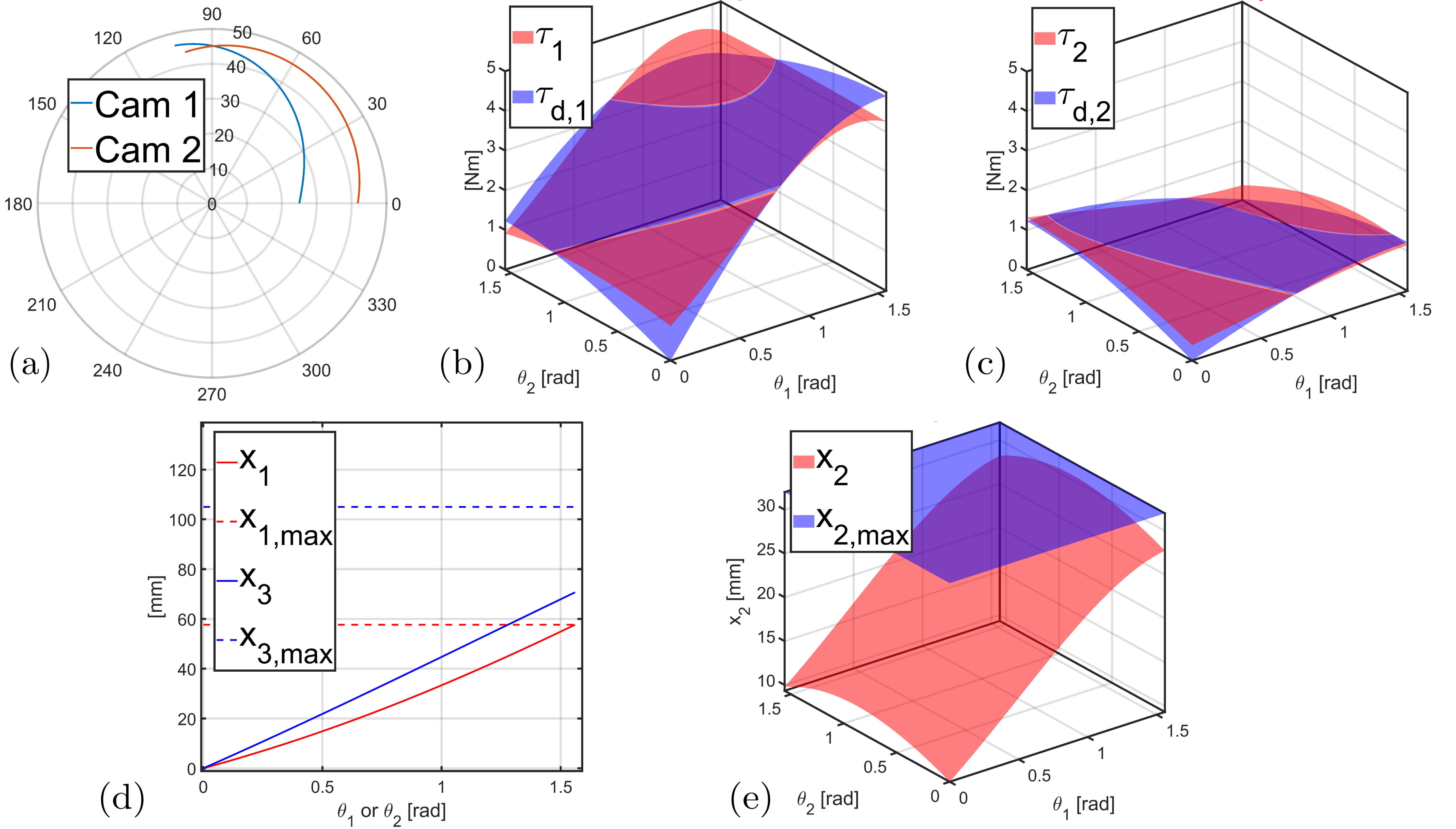}
    \caption{Two DOF simulation results without sensitivity minimization: (a) cam profiles (b) $\tau_1$ vs. $\tau_{d,1}$ (c) $\tau_2$ vs. $\tau_{d,2}$ (d) $x_1$, $x_{1,max}$, $x_3$, and $x_{3,max}$ (e) $x_2$ and $x_{2,max}$}
    \label{fig:sim_no_sens_2DOF}
\end{figure*}
For this simulation, the scalar weights were chosen to be $w_1 = w_2 = 10$ and $w_3 = \dots = w_8 = 0$. Setting $w_3$ through $w_8$ to zero makes optimizer not attempt to minimize the sensitivity to changes in spring constant. Figure \ref{fig:sim_no_sens_2DOF} shows the results of the simulation. Figure \ref{fig:sim_no_sens_2DOF}(a) shows a polar plot of the cam designs; Fig. \ref{fig:sim_no_sens_2DOF}(b) shows the difference between the desired cam 1 torque $\tau_{d,1}$ and the actual cam torque returned by the optimizer $\tau_1$; Fig. \ref{fig:sim_no_sens_2DOF}(c) shows the difference between the desired cam 2 torque $\tau_{d,2}$ and the actual cam torque returned by the optimizer $\tau_2$; Fig. \ref{fig:sim_no_sens_2DOF}(d) shows the deflection of springs 1 and 3 as a function of $\theta_1$ and $\theta_2$, respectively. Additionally, the plot shows the maximum allowable deflections in the springs; Fig. \ref{fig:sim_no_sens_2DOF}(e) shows the deflection in spring 2 as a function of $\theta_1$ and $\theta_2$. It also shows the maximum spring deflection. Table \ref{tab:no_sens_results} shows the cam polynomial coefficients and spring pre-extensions returned by the optimization routine. Also, Table \ref{tab:no_sens_results} shows the \cut{root mean square error (}RMSE\cut{)} and maximum error between $\tau_1$ and $\tau_{d,1}$ and between $\tau_2$ and $\tau_{d,2}$.\corrlab{R1-3}{ Our implementation generated these results in 8.60 minutes.} 
\begin{table}[htbp]
\renewcommand{\arraystretch}{1.0}
\caption{Simulation results without minimizing sensitivity}\label{tab:no_sens_results}
\fontsize{8pt}{10pt}\selectfont  
\begin{center}
\small
\begin{tabular}{ |c|c| }
\hline
 \textbf{Parameter} & \textbf{Value}\\ \hline
 $\bs{\beta}_1$ & $[-0.0052, 0.0133, 0.0046, 0.0250]\T$ \\ \hline
 $\bs{\beta}_2$ & $[-0.0009, -0.0016, 0.0068, 0.0417]\T$  \\ \hline
 $\mb{x}_0$     & $[0, 9.33, 0]$ mm  \\ \hline
 $\tau_1$ vs. $\tau_{d,1}$ RMSE & 243.12 Nmm \\ \hline
 $\tau_1$ vs. $\tau_{d,1}$ Max. error & 868.25 Nmm \\ \hline
 $\tau_2$ vs. $\tau_{d,2}$ RMSE & 124.04 Nmm \\ \hline
 $\tau_1$ vs. $\tau_{d,2}$ Max. error & 389.92 Nmm \\ \hline
\end{tabular}
\end{center}
\end{table}
\subsection{Two DOF System With Minimizing Sensitivity}
\begin{figure*}[htbp]
    \centering
    \includegraphics[width=0.9\textwidth]{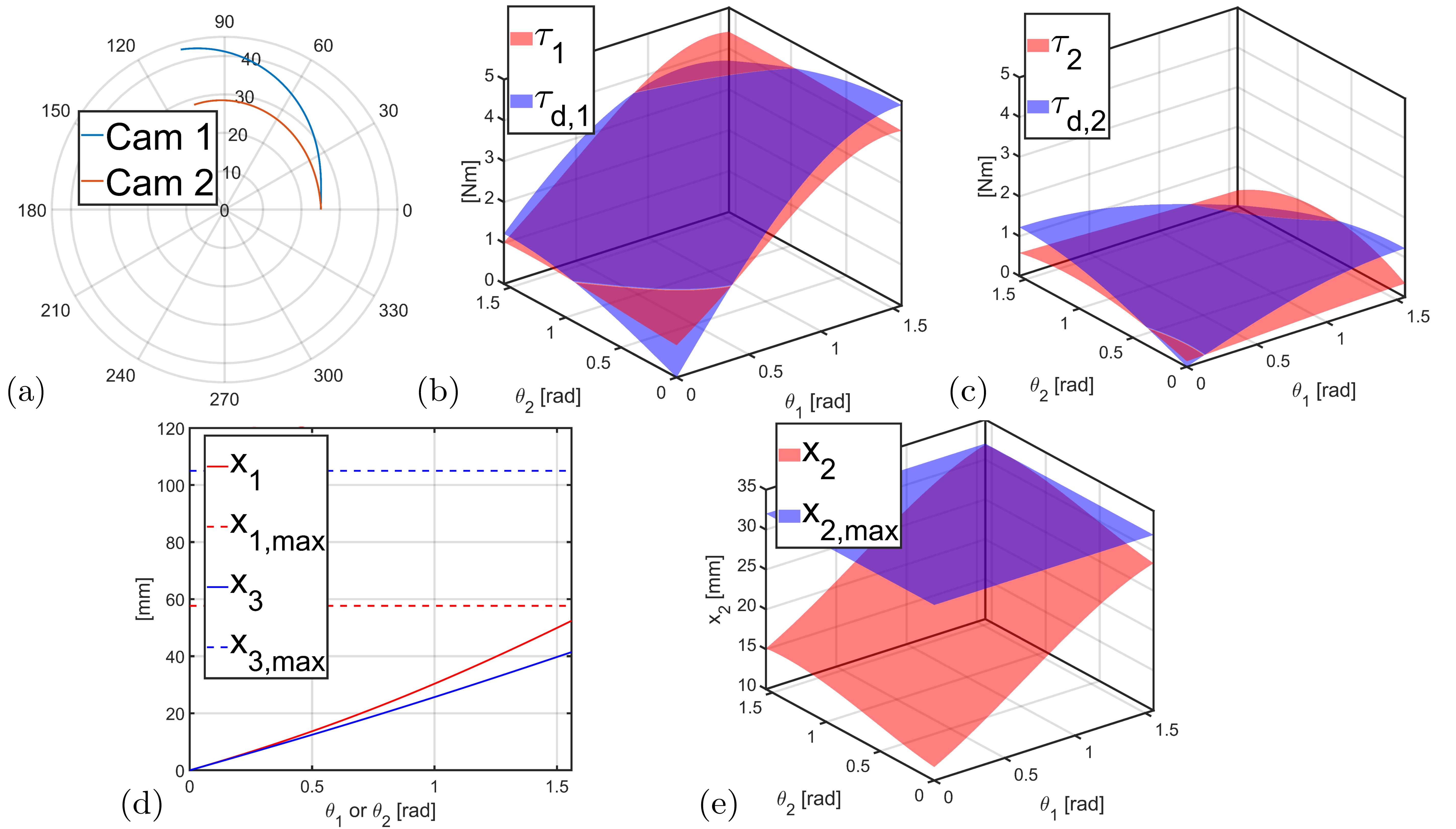}
    \caption{Two DOF simulation results with sensitivity minimization: (a) cam profiles (b) $\tau_1$ vs. $\tau_{d,1}$ (c) $\tau_2$ vs. $\tau_{d,2}$ (d) $x_1$, $x_{1,max}$, $x_3$, and $x_{3,max}$ (e) $x_2$ and $x_{2,max}$}
    \label{fig:sim_w_sens_2DOF}
\end{figure*}
For this simulation, the scalar weights were chosen to be $w_1 = w_2 = 1$ and $w_3 = \dots = w_8 = 1000$. This will cause the optimizer to minimize the sensitivity to changes in spring constant. These values were chosen so that the terms of Eq.~\eqref{eq:opt_2dof} have similar magnitudes. Figure \ref{fig:sim_w_sens_2DOF} shows the results of the simulation. Figure \ref{fig:sim_w_sens_2DOF}(a) shows a polar plot of the cam designs; Fig. \ref{fig:sim_w_sens_2DOF}(b) shows the difference between the desired cam 1 torque $\tau_{d,1}$ and the actual cam torque returned by the optimizer $\tau_1$; Fig. \ref{fig:sim_w_sens_2DOF}(c) shows the difference between the desired cam 2 torque $\tau_{d,2}$ and the actual cam torque returned by the optimizer $\tau_2$; Fig. \ref{fig:sim_w_sens_2DOF}(d) shows the deflection of springs 1 and 3 as a function of $\theta_1$ and $\theta_2$, respectively. Additionally, the plot shows the maximum allowable deflections in the springs; Fig. \ref{fig:sim_w_sens_2DOF}(e) shows the deflection in spring 2 as a function of $\theta_1$ and $\theta_2$. It also shows the maximum spring deflection. Table \ref{tab:sens_results} shows the cam polynomial coefficients and spring pre-extensions returned by the optimization routine. Also, Table \ref{tab:sens_results} shows the RMSE and maximum error between $\tau_1$ and $\tau_{d,1}$ and between $\tau_2$ and $\tau_{d,2}$. \corrlab{R1-3}{Our implementation generated these results in 5.67 minutes.} 
\begin{table}[htbp]
\renewcommand{\arraystretch}{1.0}
\fontsize{8pt}{10pt}\selectfont  
\caption{Simulation results with minimizing sensitivity}\label{tab:sens_results}
\begin{center}
\begin{tabular}{ |c|c| }
\hline
 \textbf{Parameter} & \textbf{Value} \\ \hline
 $\bs{\beta}_1$ & $[-0.0041, 0.0125, 0.0007, 0.0250]\T$  \\ \hline
 $\bs{\beta}_2$ & $[-0.0019, 0.0052, -0.0014, 0.0251]\T$  \\ \hline
 $\mb{x}_0$     & $[0, 11.68, 0]\T$ mm \\ \hline
 $\tau_1$ vs. $\tau_{d,1}$ RMSE & 415.00 Nmm \\ \hline
 $\tau_1$ vs. $\tau_{d,1}$ Max. error & 788.33 Nmm \\ \hline
 $\tau_2$ vs. $\tau_{d,2}$ RMSE & 384.84 Nmm \\ \hline
 $\tau_2$ vs. $\tau_{d,2}$ Max. error & 428.17 Nmm \\ \hline
\end{tabular}
\end{center}
\end{table}
\subsection{Sensitivity Minimization Results}
To demonstrate that the method indeed minimizes the sensitivity of the cam torques, the spring constants shown in Table \ref{tab:sim_consts} were increased by \corrlab{R1-6}{20\%, 10\%, and 5\%} and the torque on both cam designs was calculated with the increased spring constants. The RMSEs between the original cam torques and the cam torques with increased spring constants were calculated and the results are shown in Table \ref{tab:sens_min_results}. From these results, it is clear that the deviation in cam torque due to unexpected changes in spring constant is reduced by the optimization routine.

\begin{table*}[htbp]
\renewcommand{\arraystretch}{1.0}
\caption{\corrlab{R1-6}{Deviation in cam torque after increase in spring constants}}\label{tab:sens_min_results}
\fontsize{8pt}{10pt}\selectfont  
\small
\begin{center}
\begin{tabular}{ |c|c|c|c|c| }
\hline
  {}&{}&\textbf{Without Sens. Min. (Fig. \ref{fig:sim_no_sens_2DOF})} & \textbf{With Sens. Min. (Fig. \ref{fig:sim_w_sens_2DOF})} & \textbf{Reduction in RMSE}\\ \hline
 \multirow{2}{*}{20\% increase} & Cam 1 RMSE & 702.76 Nmm & 647.05 Nmm & 55.71 Nmm   \\ 
                                & Cam 2 RMSE & 204.69 Nmm & 143.84 Nmm & 60.85 Nmm \\ \hline
 \multirow{2}{*}{10\% increase} & Cam 1 RMSE & 351.38 Nmm & 323.52 Nmm & 27.86 Nmm   \\ 
                                & Cam 2 RMSE & 102.34 Nmm &  71.92 Nmm & 30.42 Nmm \\ \hline
  \multirow{2}{*}{5\% increase} & Cam 1 RMSE & 175.69 Nmm & 161.76 Nmm & 13.93 Nmm   \\ 
                                & Cam 2 RMSE &  51.17 Nmm & 35.961 Nmm & 15.21 Nmm \\ \hline
\end{tabular}
\end{center}
\end{table*}
\section{Experimental Verification}\label{sec:experiment}
\par
\begin{figure}[htbp]
    \centering
    \includegraphics[width=0.9\columnwidth]{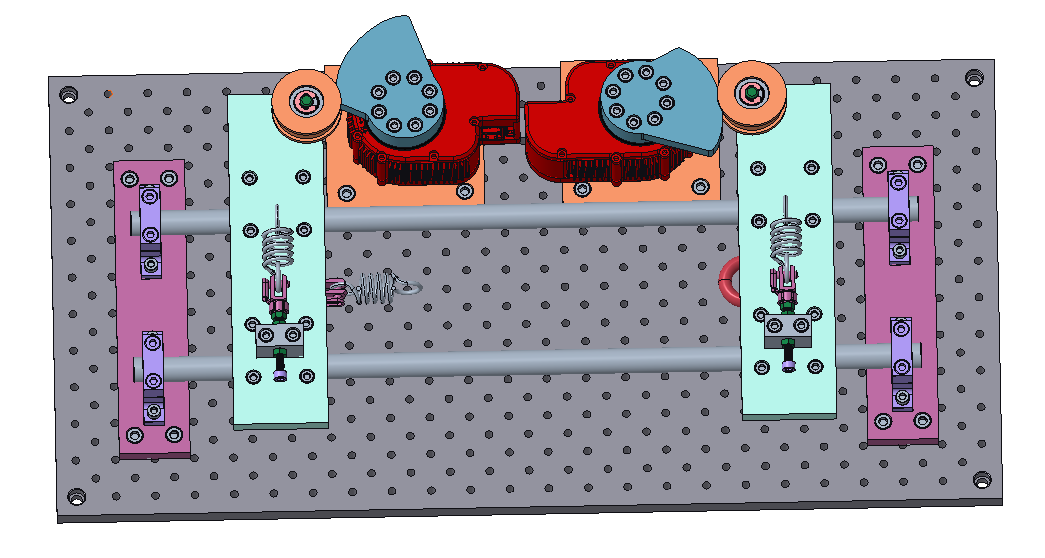}
    \caption{Experimental setup: \protect\circled{1} cam 1, \protect\circled{2} cam 2, \protect\circled{3} idler 1, \protect\circled{4} idler 2, \protect\circled{5} spring 1, \protect\circled{6} spring 2, \protect\circled{7} spring 3, \protect\circled{8} Hebi\texttrademark~X8-16 actuator, \protect\circled{9} linear shaft, \protect\circled{10} linear ball bearing}
    \label{fig:setup_2dof}
\end{figure}

 To experimentally validate our model of the cam torques, we manufactured a desktop prototype of the 2 DOF cam system as shown in Fig. \ref{fig:setup_2dof} \corrlab{R1-10}{using the springs listed in section \ref{subsec:sim_IC}}. The setup consists of 3D printed cams (Fig. \ref{fig:setup_2dof}\circled{1} and \circled{2}) attached to Hebi\texttrademark~X8-16 actuators (Fig. \ref{fig:setup_2dof}\circled{8}). The cams roll against bearing mounted, 3D printed idlers (Fig. \ref{fig:setup_2dof}\circled{3} and \circled{4}). The idlers are mounted on linear bearings (Fig. \ref{fig:setup_2dof}\circled{10}) between which spring 2 is mounted (Fig. \ref{fig:setup_2dof}\circled{6}). The wire ropes are attached to the cams on one end, pass through grooves in the idlers, and are terminated on springs 1 (Fig. \ref{fig:setup_2dof}\circled{5}) and 3 (Fig. \ref{fig:setup_2dof}\circled{7}).
\par The torque on the cams are measured using the deflection in the series-elastic elements on the Hebi\texttrademark~actuators. The spring constants of the series-elastic elements were calibrated by applying known moments to the actuators, recording the deflections, and finding the spring constants that best fit the data. \corrlab{R1-7}{The calibrated spring constant for the Hebi\texttrademark~actuator attached to cam 1 was 337.83 Nm/rad and the calibrated spring constant of the Hebi attached to cam 2 was 148.68 Nm/rad.} 
\par According to the manufacturer, spring 2 has 23.22 N of pre-tension. However, the spring force given in Eq.~\eqref{eq:f2} assumes there is no pre-tension in the spring. To account for this, the pre-extension in spring 2 given in Table \ref{tab:sens_results} was lowered to 8.52 mm so that the force on the cams from spring 2 when $\theta_1 = \theta_2 = 0$ is equal to the force that was expected using Eq.~\eqref{eq:f2}.
\par
\begin{figure*}[h]
    \centering
    \includegraphics[width=0.85\textwidth]{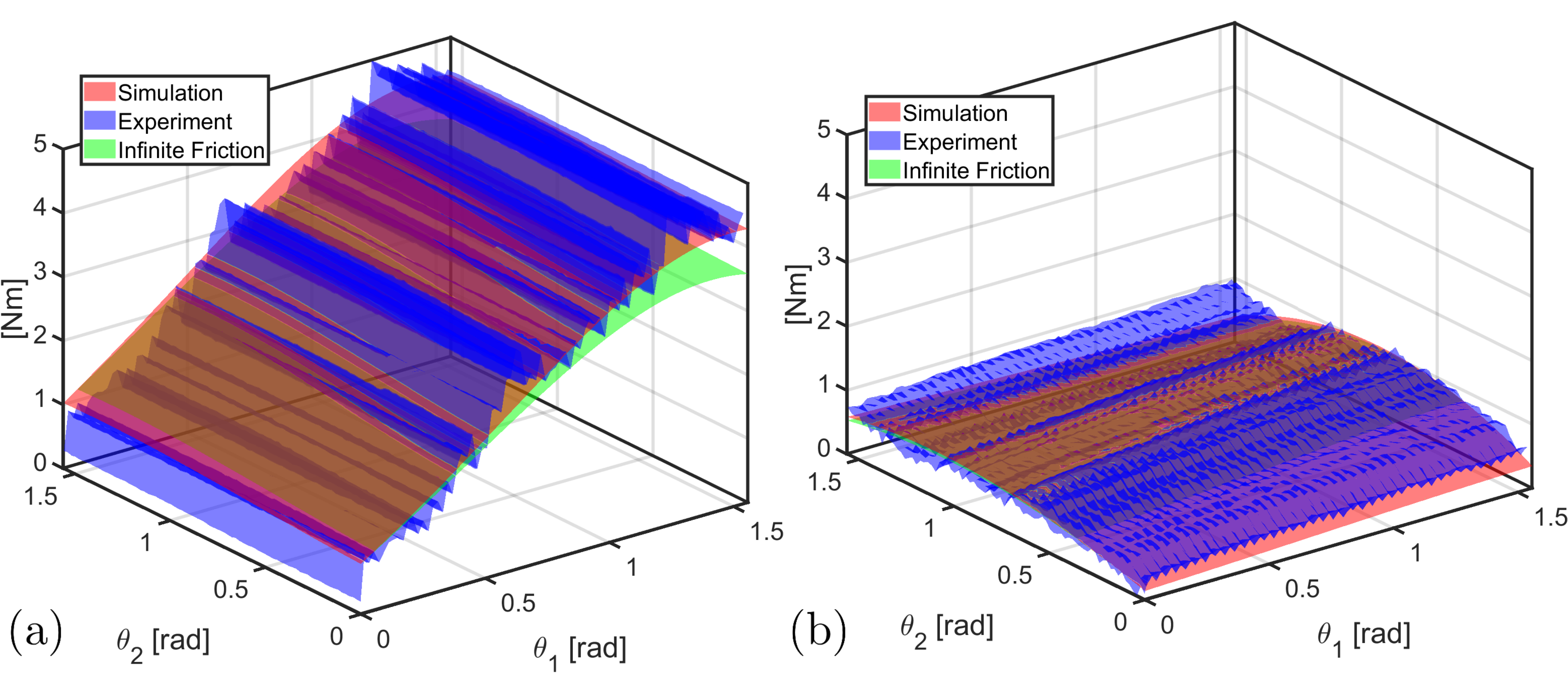}
    \caption{Experimental cam torques compared to simulated cam torques with friction model and simulated cam torques with infinite friction: (a) cam 1 (b) cam 2}
    \label{fig:results_2dof}
\end{figure*}

 \remind{R3-4}{As shown in this paper's \href{https://youtu.be/tqnUBkm_f_M}{multimedia extension}, both actuators start at $\theta_1=\theta_2 = 0^\circ$. Cam 1 is kept fixed and cam 2 sweeps from $0^\circ$ to $ 90^\circ$ in 0.02 rad increments. At each increment, 20 torque measurements are taken on each cam. When $\theta_2 = 90^\circ$, $\theta_1$ is incremented by 0.02 rad and then $\theta_2$ decreases to $0^\circ$ by 0.02 rad increments, taking 20 measurements at each increment. This process is repeated until $\theta_1 = 90^\circ$.} The results of this experiment are shown in Fig. \ref{fig:results_2dof}. In this test, the RMSE between the model predicted torque and the actual torque for cam 1 was 353.0 Nmm. \remind{R2-2}{However, when friction was ignored, the RMSE increased to 518.6 Nmm.} The RMSE between the model predicted torque and the actual torque for cam 2 was 166.3 Nmm. \remind{R2-2}{With infinite friction, the RMSE was 174.1 Nmm.} \corrlab{R1-8}{The RMSE between the actual torque and $\bs{\tau}_{d,1}$ was 579.9 Nmm and 356.2 Nmm between the actual torque and $\bs{\tau}_{d,2}$.} 
\par \remind{R2-1}{Potential sources of error in this experiment include the accuracy and noise of the torque sensing on the Hebi actuators, friction in the linear bearings, the accuracy of the friction characterization experiment from section \ref{sec:fric_char}, \cut{and} inexact pre-extension adjustment}, \corrlab{R2-2}{and deflection}\corrlab{R3-4}{/hysteresis in the wire-rope}. \corrlab{R1-10}{Additionally, the stiffness of springs 1-3 were assumed to be equal to the manufacturer-specified value. This assumption introduced additional error between the experimental and model-predicted torques.}
\section{Discussion}\label{sec:disc}
\par While this method is guaranteed to return a physically realizable cam system, there is no guarantee that the resulting cam torques  will exactly match the desired torque. We believe this is acceptable for robotic applications where the actuators are able to compensate for differences between the static torques and the torques provided by the cams. \remind{R2-5}{For discontinuous or highly non-monotonic desired torque profiles,  the torque matching performance will degrade. Additionally, if the desired torque on one the cams is a stronger function of the other cam angle than of its own angle, the optimization routine may struggle to match the desired torque profile. However, some of this can be alleviated by increasing $|a_0|$ (i.e. increasing the moment arm of spring 2's force). Additionally, if the constraints on the cam design (e.g., $\rho_{min},~\rho_{max}$, etc.) are too narrow, the desired and actual torques may not match well.} \corrlab{R2-5}{Additionally, this cam design only works for scenarios when the distance between the cam centers of rotation remains constant.}
\par \remind{R3-2}{Because the partial derivative of the cam torques with respect to the spring constants (Eqns.~\eqref{eq:dtau_dk} and \eqref{eq:tau1_sens_1dof}) are strong functions of the spring deflections and cam-idler contact point locations, minimizing the sensitivity to unexpected changes in spring constant is essentially equivalent to minimizing the size of the cams. This means that increasing values of $w_3$ through $w_8$ will sometimes decrease how well the torque matches the desired torque.}
\par Lastly, the friction model presented in Section \ref{sec:finite_friction} has less impact on the torque for relatively circular cams. This is because the distributed forces that are locally perpendicular to the cam $\mb{f}_c$ pass through the cam center of rotation and therefore cannot create a moment about the cam center of rotation. This can be seen in the experimental results presented in Section \ref{sec:experiment}. In Fig. \ref{fig:results_2dof}a, the difference between the finite and infinite torque curves is much larger than in Fig. \ref{fig:results_2dof}b. This is because cam 2 is much closer to circular than cam 1, which can be seen in Figs. \ref{fig:sim_w_sens_2DOF}a and \ref{fig:setup_2dof}.    
\section{Conclusions}\label{sec:conclusion}
\par Collaborative robots must simultaneously pose minimal risk to workers and be powerful enough to assist workers in industrial tasks such as lifting heavy equipment. One method for navigating these conflicting design requirements is through the use of static balancing mechanisms to offset the robot's self-weight, thus enabling the selection of lower powered (i.e. safer) actuators.
\par Because of their simple, lightweight, and compact design, wire-wrapped cam mechanisms are a promising option for static balancing. However, previous works on wire-wrapped cam mechanisms can return non-convex cams, require unrealistically long spring torques, and ignore the effect of friction between the wire and the cam. These methods are also sensitive to unmodeled deviations in spring constant.
\par To address these limitations, in this paper, we presented the design of a novel, two DOF wire-wrapped cam system where the torque on each cam is a function of both cam angles. We also presented a model of how the friction and distributed pressure between the cam and the wire affects the torque on the cams. This friction model takes into account the distributed friction and contact pressure between the wire and the cam. This relaxes key assumptions made in previous works that treated the torque on the cam as resulting from a single point force at the wire tangency point. Using this model, we presented an optimization-based cam design procedure  that 1) ensured the cam is convex, 2) guaranteed the spring deflections stay below the maximum allowable values, and 3) minimized sensitivity to unexpected changes in spring constant.
\par Using this cam design method, we built a prototype of the proposed mechanism and experimentally determined that our model predicts the torque on the cams to within 353.0 Nmm of RMS error. The results also indicate that the distributed force between the wire and the cam can have a significant effect on cam torque for more non-circular cams. 

\section{Multimedia Extension}
The multimedia extension for this paper can be found at \url{https://youtu.be/tqnUBkm_f_M}. 
\section*{Acknowledgement} 
This work was supported by NSF award \#1734461 and by Vanderbilt internal university funds. A. Orekhov was partially supported by the NSF Graduate Research Fellowship under \#DGE-1445197.
 

\bibliographystyle{asmems4}
\bibliography{bib/garrison_refs,bib/nabil_refs,bib/andrew_refs}
\end{document}